\documentclass[11pt]{article}

\usepackage[margin=1in]{geometry}
\usepackage{graphicx}
\usepackage{booktabs}
\usepackage{amsmath,amssymb}
\usepackage[T1]{fontenc}
\usepackage[authoryear,round]{natbib}
\usepackage{titlesec}
\usepackage{changepage}
\usepackage{url}
\usepackage{algorithm}
\usepackage{algpseudocode}
\usepackage{etoolbox}
\usepackage{tabularx}        %
\usepackage{array}           %
\usepackage{xurl}            %
\usepackage{fancyvrb}        %
\usepackage{subcaption}
\usepackage{wrapfig}         
\usepackage{minted}
\usepackage{booktabs}
\usepackage{longtable}

\newcolumntype{L}{>{\raggedright\arraybackslash}X}

\newcommand{\sentbench}{\texttt{SentinelBench}\xspace}

\IfFileExists{newtxtext.sty}{%
  \usepackage{newtxtext,newtxmath}%
}{%
  \usepackage{mathptmx}%
}

\IfFileExists{matheus-style.sty}{%
  \usepackage{matheus-style}%
}{%
  \IfFileExists{matheus-style.tex}{\input{matheus-style.tex}}{}%
}

\usepackage[
  colorlinks=true,
  linkcolor=blue,
  citecolor=blue,
  urlcolor=blue
]{hyperref}

\usepackage{xcolor}

\makeatletter
\def\@maketitle{%
  \newpage
  \null
  \vspace*{-0.5in}
  \begin{flushleft}
    {\noindent\resizebox{\linewidth}{!}{\scshape\@title}\par}
    \vspace{0.9em}
    {\normalsize\lineskip .35em \@author\par}
  \end{flushleft}
  \vspace{0.1em}
}
\makeatother

\renewenvironment{abstract}
  {%
    \vspace{0.3em}%
    \begin{center}\scshape ABSTRACT\end{center}
    \begin{adjustwidth}{1.2cm}{1.2cm}%
      \small\noindent\ignorespaces
  }
  {%
    \end{adjustwidth}%
    \vspace{0.4em}%
  }

\titleformat{\section}
  {\normalfont\Large\scshape}
  {\thesection}
  {0.75em}
  {}
\titleformat{\subsection}
  {\normalfont\large\scshape}
  {\thesubsection}
  {0.7em}
  {}
\titleformat{\subsubsection}
  {\normalfont\normalsize\scshape}
  {\thesubsubsection}
  {0.6em}
  {}
\titlespacing*{\subsubsection}{0pt}{0.6em}{0.25em}
\titlespacing*{\subsection}{0pt}{0.75em}{0.3em}
\titlespacing*{\section}{0pt}{1.0em}{0.45em}

\pretocmd{\appendix}{%
  \titleformat{\section}
    {\normalfont\LARGE\scshape}
    {\thesection}
    {0.75em}
    {}
  \titleformat{\subsection}
    {\normalfont\large\scshape}
    {\thesubsection}
    {0.7em}
    {}
}{}{}

\title{SentinelBench: A Benchmark for Long-Running Monitoring Agents}
\author{
\hspace*{0.35em}\begin{minipage}{0.95\textwidth}
\textbf{Matheus Kunzler Maldaner}$^{1}$ \hspace{0.3cm}
\textbf{Adam Fourney}$^{2}$ \hspace{0.3cm} 
\textbf{Amanda Swearngin}$^{2}$ \hspace{0.3cm}
\textbf{Hussein Mozannar}$^{2}$ \hspace{0.3cm}
\textbf{Gagan Bansal}$^{2}$ \hspace{0.3cm}
\textbf{Maya Murad}$^{2}$ \hspace{0.3cm}
\textbf{Rafah Hosn}$^{2}$ \hspace{0.3cm}
\textbf{Saleema Amershi}$^{2}$
\end{minipage}\\[10pt]
\hspace*{0.35em}$^{1}$University of Florida \hspace{0.6cm} $^{2}$Microsoft Research, AI Frontiers
}
\date{}

\begin{document}
\maketitle

\begin{abstract}
AI agents are increasingly asked to carry out work that spans minutes, hours, or longer. Yet the default model of agent behavior is continuous action: issuing tool calls, refreshing pages, searching for alternatives, or otherwise trying to force progress. This is the wrong approach for many long-running tasks, which are better served by a strategy of sustained attention. Instead, agents should monitor an environment, notice when an external event makes progress possible, then respond promptly without wasting resources while waiting. To measure progress on this class of tasks, we introduce \sentbench, an open-source benchmark for time-evolving monitoring tasks.

\sentbench contains 100 tasks across 10 synthetic web environments, including email, calendars, finance, professional networking, and entertainment. Each environment exposes a live web interface and replays a scripted sequence of events, requiring agents to navigate and reason about web pages whose state shifts underfoot. \sentbench measures task completion, reaction time, and resource use, exposing the tradeoff between responsiveness and cost. We report results across three models and two browser-agent harnesses, establishing performance baselines for future comparison and demonstrating how agent design choices can dramatically impact key metrics. Together, these results show that \sentbench distinguishes meaningful differences in agent behavior.
\end{abstract}

\begin{figure}[h]
    \centering
    \includegraphics[width=0.85\linewidth]{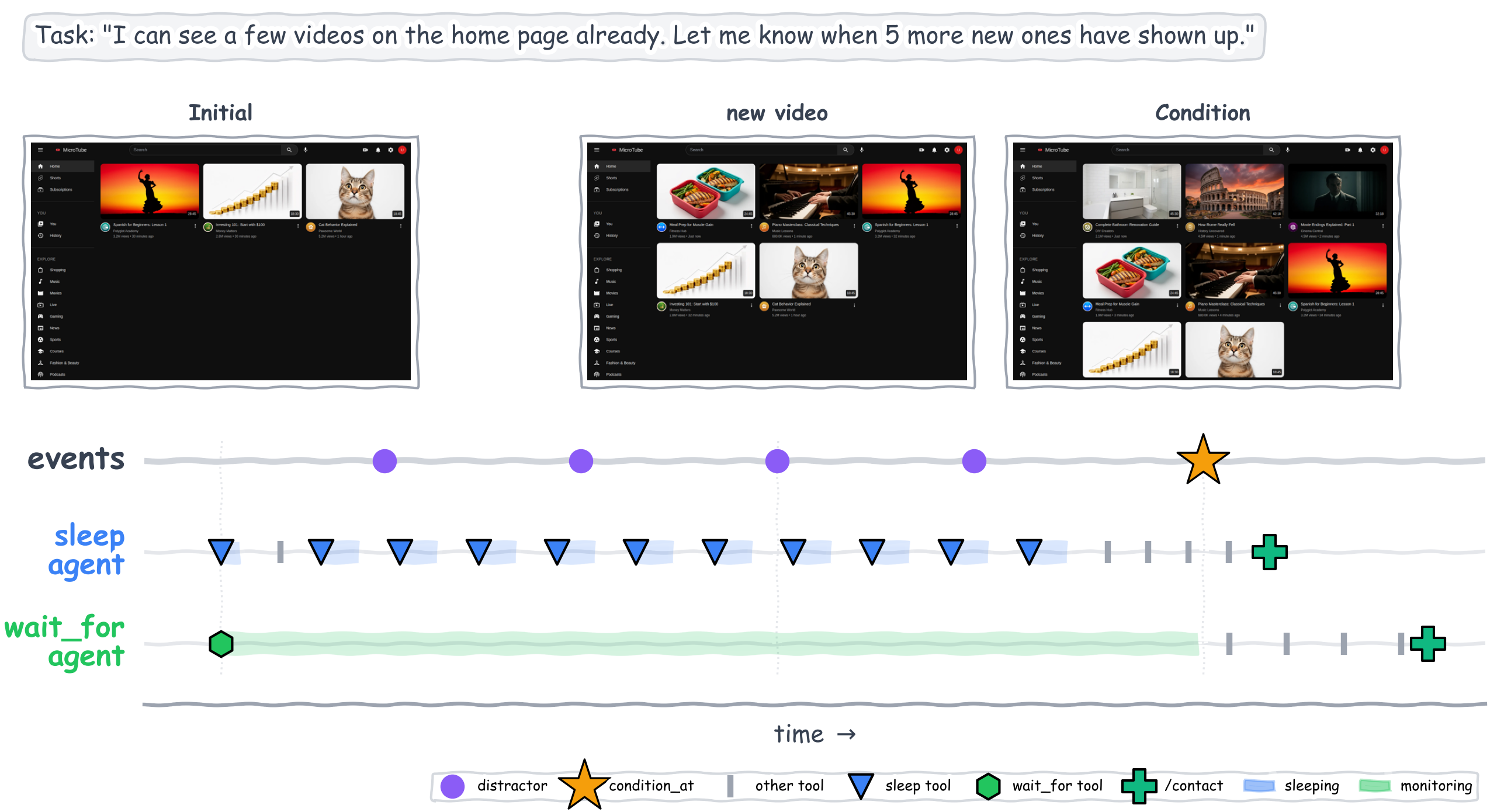}
  \caption{Agent execution timelines in \sentbench for a representative scenario. Each row visualizes one agent strategy over simulated time as events are played back in the simulation (screenshots show the state of the app after selected events). The \texttt{sleep} agent spends most of its time executing fixed interval polling while the \texttt{wait\_for} agent waits for a condition for an extended period and resumes once the environment changes to meet the condition. \sentbench enables systematic investigation of agent implementation choices for measuring progress on time-evolving monitoring tasks.}
    \label{fig:teaser}
\end{figure}

\begin{figure}[b]
    \centering
    \includegraphics[width=0.98\linewidth]{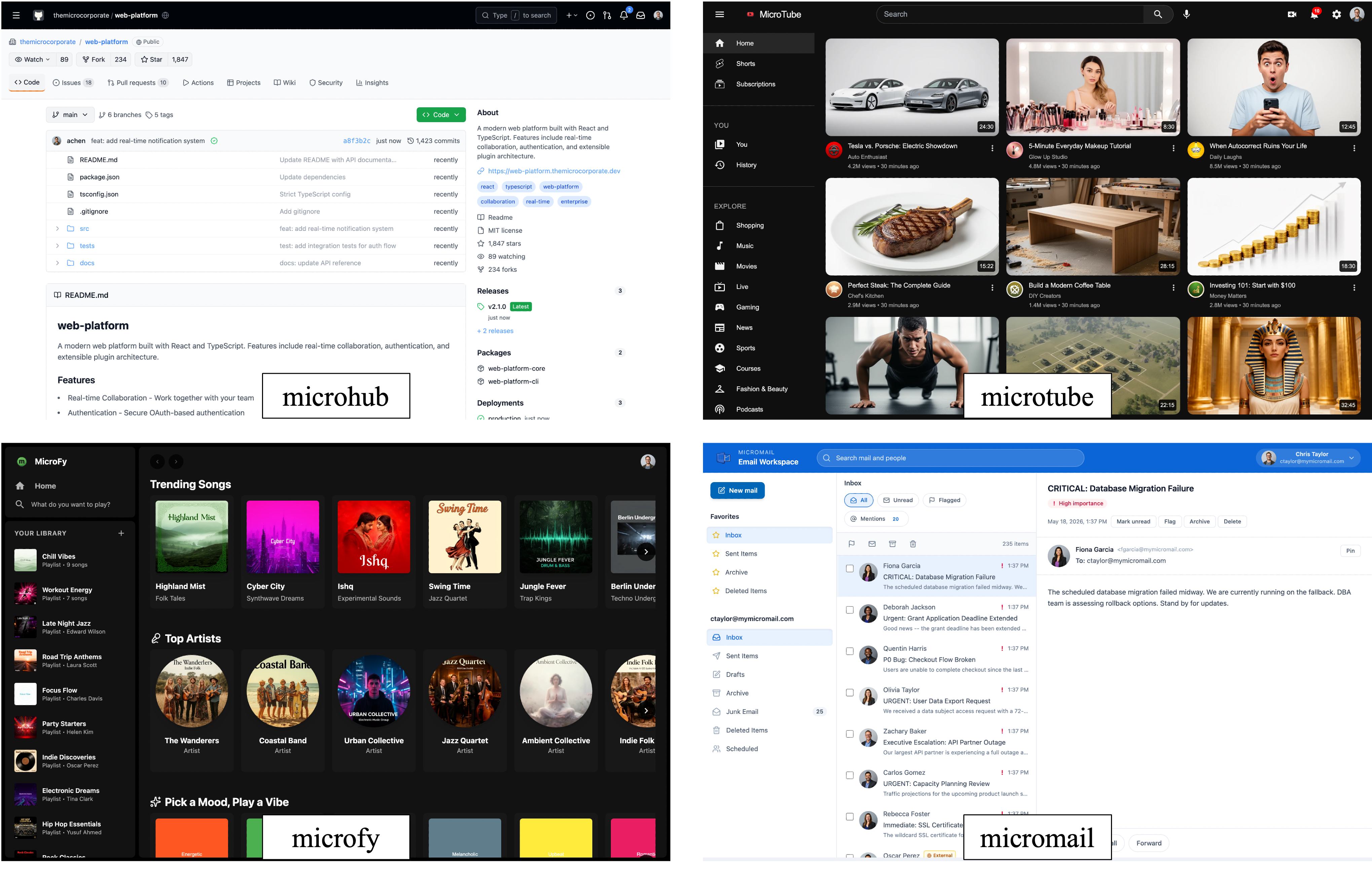}
    \caption{Four \sentbench environments. Each consists of a synthetic environment with a live web interface and a set of monitoring tasks with scripted sequences of events that can be played back to simulate the environment evolving over time.}
    \label{fig:environments}
\end{figure}




\section{Introduction}
\label{sec:introduction}

Modern AI agents are increasingly capable of long and complex work. METR, for example, has tracked the 50\% task-completion time horizon of AI models and found that it has doubled approximately every seven months, increasing from 4 seconds in 2019 to more than 16 hours in 2026 \cite{metr2025domains, kwa2025metr}. In practice, developers already use coding agents for long-running work such as large codebase migrations and comprehensive research projects that take hours to complete \cite{anthropic2025contextengineering}.

While impressive, these workflows often embed an implicit assumption: that meaningful state changes in the environment result from the agent’s direct actions, such as executing tools or communicating with other agents. In other words, if the environment is not in a state enabling forward progress, the agent needs to move the environment toward such a state. This, of course, is not always a correct model of the world. Many tasks require an agent not to act continuously, but instead to watch and wait until conditions become conducive to task completion. For example, no amount of reloading a webpage will make concert tickets go on sale faster, nor will broadening a web search help when ticket sales are exclusive to a particular site. When agents are compelled to always take action, they waste resources at best, and fail the task outright at worst, especially since task success often drops precipitously as trajectory length increases \cite{sinha2026illusion}.

In previous work, we presented designs for agents that are better able to handle these monitoring-like tasks \cite{mozannar2025magenticui}. We referred to such tasks as \emph{Sentinel Tasks} and previewed a benchmark for measuring agent performance on these workloads \cite{mozannar2025sentinel}. In \emph{this} paper, we formally present and release that benchmark, which we call \sentbench (Figure~\ref{fig:environments}). The full benchmark (code, environments, task scenarios, synthetic catalogs, and data-generation pipeline) is publicly made available at \url{https://github.com/microsoft/sentinel_environments}.

\sentbench is a collection of 100 tasks divided across 10 synthetic web environments, including email, calendars, finance, professional networking, entertainment, and others. Each task scenario replays a sequence of events over time and asks the agent to complete work in this live, self-evolving setting. For example, one task asks the agent to ``\emph{Keep an eye on the jobs page. If a new role mentions Kubernetes in the requirements, apply to it and let me know.}'' The agent succeeds if it notices and applies to the job posting, and then takes steps to contact the user. The benchmark also measures \emph{reaction time} (how soon after the condition was met the agent completed the task) and resource utilization\footnote{While \sentbench is designed to work with any web browsing agent, token-based metrics require a cooperative agent that self-reports token use at the end of the task.} (the number of tokens consumed and generated). For many strategies, such as polling, this sets up a natural tension: poll too frequently and costs soar; poll too slowly, and reaction time rises. By default, each of the 100 tasks is designed to be achievable within 10 minutes, but a \emph{speed\_factor} can be applied to stretch tasks to much longer durations making the reaction-cost tradeoff even more stark.

To demonstrate that \sentbench can meaningfully measure a broad range of outcomes across model scales and agent capabilties, we evaluate three models and two agent harness configurations against \sentbench. Specifically, we consider GPT-5.4 (low reasoning) to represent a newer agentic frontier model, Qwen 3.5:9b to represent a local agentic model, and GPT-4o to represent an older frontier chat model. We pair each model with a web browsing harness that either includes a classic \texttt{sleep(time)} tool, which allows an agent to unconditionally yield for a specified amount of time, or a purpose-built \texttt{wait\_for(condition, timeout)} tool, which allows agents to heuristically detect and respond to changes in page content. Across these data points, with the default speed\_factor, we observe task completion rates ranging from 46\% to 75\%, while token usage ranges from $70,000$ to over $500,000$ per task (Table~\ref{tab:success_rate}). When tasks are extended to 40 minutes, we show that GPT-5.4 agents using \texttt{sleep(time)} can cost $10x$ more ($\$4.65$ vs $\$0.48$ USD), while completing fewer total tasks ($56\%$ vs $69\%$). This demonstrates both that \sentbench can measure these important metrics and that design choices can dramatically affect the viability of an agent in performing these tasks. 

In the remainder of this paper we detail the 10 environments, and characterize the tasks and key elements of their design. We then review the task life cycle, the protocols driving environment events, and the metrics computed by the benchmark. Next, we present detailed findings from the evaluations described above, demonstrating both the need to measure agents on these tasks and that \sentbench is fit for this purpose. Finally, we compare \sentbench to related work and discuss future evolutions of this work.

\begin{figure}[t]
    \centering
    \begin{minipage}[t]{0.54\linewidth}
        \vspace{0pt}       
  {\small                                                                                 
  \begin{tabular}{@{}>{\ttfamily\bfseries}r @{:\hspace{6pt}} p{0.78\linewidth}@{}}                         
  id              & microfy-lyric-subway-absolute-active \\                                                
  environment     & microfy \\                              
  start\_page     & \texttt{HOST\_ADDRESS/microfy} \\
  prompt          & ``Watch the trending feed. When a song drops whose lyrics mention `subway', like it for
  me.'' \\
  condition\_at   & 329.11\,s \\
  kill\_at        & 630.0\,s \\
  timeline\_end   & 720.0\,s \\[4pt]
events         & t\,{=}\,0.0, \texttt{preload\_tracks} (24 tracks) \\                       
  \multicolumn{1}{r}{} & t\,{=}\,119.7, \texttt{new\_track}: track-061 \\
  \multicolumn{1}{r}{} & t\,{=}\,224.4, \texttt{new\_track}: track-062 \\                   
  \multicolumn{1}{r}{} & t\,{=}\,329.1, \texttt{new\_track}: track-013\;$\bigstar$ \\       
  \multicolumn{1}{r}{} & t\,{=}\,539.2, \texttt{new\_track}: track-063 \\ [4pt]
eval\_sql       & \texttt{SELECT COUNT(*) >= 1} \\                                          
  \multicolumn{1}{r}{} & \texttt{FROM track\_states ts} \\  
  \multicolumn{1}{r}{} & \texttt{JOIN tracks t ON t.id = ts.track\_id} \\
  \multicolumn{1}{r}{} & \texttt{WHERE LOWER(t.lyrics) LIKE '\%subway\%'} \\
  \multicolumn{1}{r}{} & \texttt{AND ts.isLiked = 1} \\
  \end{tabular}}
    \end{minipage}\hfill
    \begin{minipage}[t]{0.42\linewidth}
        \vspace{0pt}
        \includegraphics[width=\linewidth]{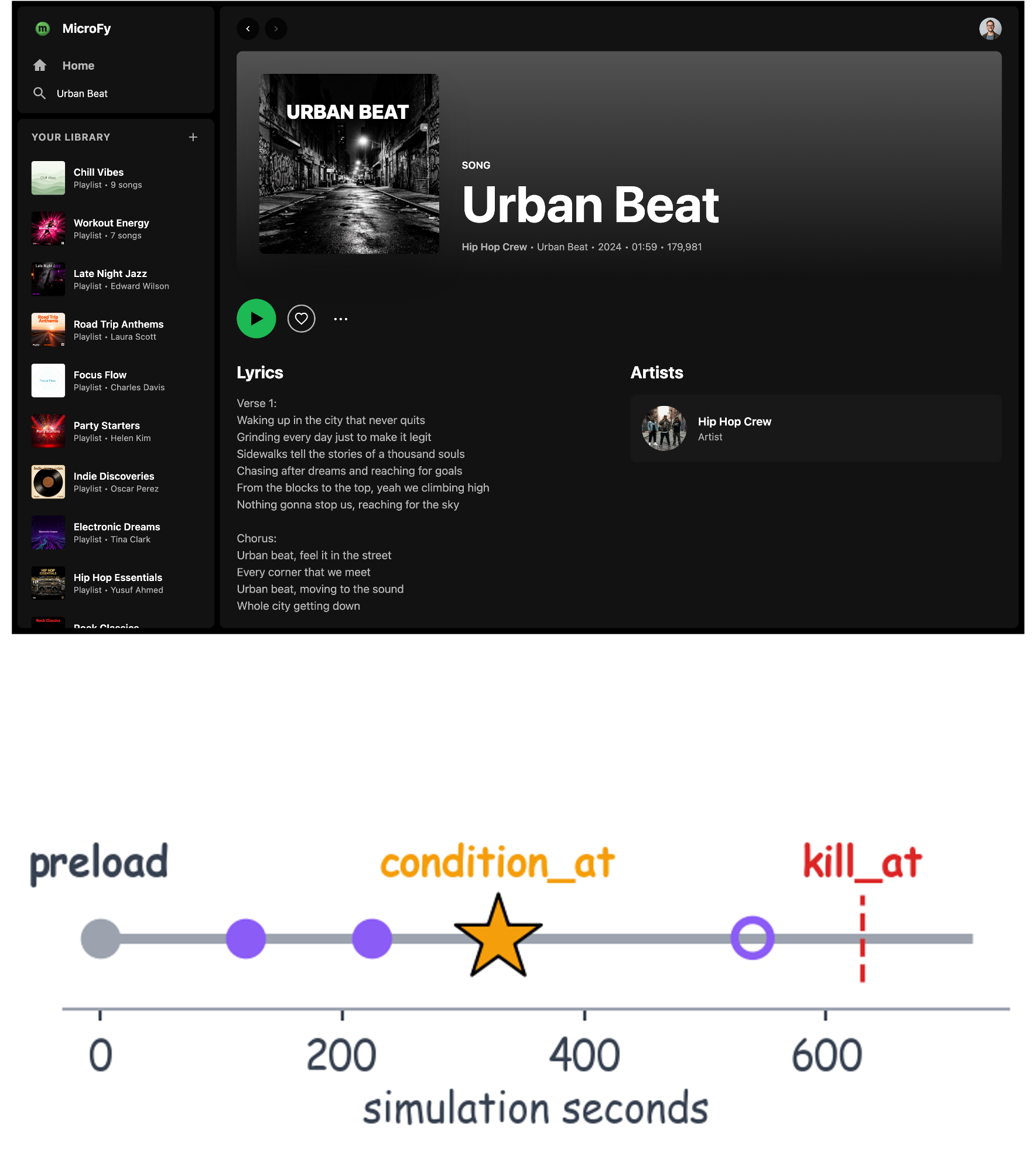}
    \end{minipage}
    \caption{Anatomy of a \sentbench monitoring task: prompt, environment UI
    at the trigger moment, event timeline, and the underlying scenario
    JSON that drives the harness (from the MicroFy environment).}
    \label{fig:sentinel_event_timeline}
\end{figure}

\section{\sentbench Environments \& Tasks}
\label{sec:env}

In this section, we provide an overview of the \sentbench environments, their design principles, and the process by which they were populated with data. We also present the tasks that make up the benchmark, and review key elements of their design.

\subsection{Overview}
\label{sec:env-overview}

\sentbench includes ten high-fidelity replicas of popular web applications as listed in Table~\ref{tab:env-overview}. Each environment is rich enough to include common interaction surfaces, and to respond in realistic manners to client actions. For example, the email application has screens for the inbox, other folders, email composition, account settings, and others. Emails are marked as read when viewed, are moved to Trash when deleted, and appear in Sent when composed and mailed. Under the hood, environments remain lightweight, modular and easy to deploy. Each is implemented as a React application served by a Vite development server, backed by a FastAPI backend and a SQLite database. These applications are built with the assumption that they are serving a single client who is already authenticated. In other words, while they present a convincing facade, they should never be deployed as, or mistaken for, production web applications.

\begin{table}[h]
\centering
\begin{tabularx}{\textwidth}{@{}l L@{}}
\toprule
\textbf{Environment} & \textbf{UI Surfaces} \\
\midrule
MicroMail    & Inbox, folders, attachments, search, flag and move actions \\
MicroChat    & Teams, channels, DMs, reactions, calls, calendars \\
MicroDin     & Feed, connections, jobs, messaging, notifications \\
MicroFy      & Tracks, playlists, artists, lyrics, follow and like actions \\
MicroGram    & Feed, stories, DMs, profiles, activity \\
MicroHood    & Portfolio, market orders, watchlist, news \\
MicroHub     & Repo, issues, PRs, commits, workflows, releases \\
MicroLendar  & Month, week, day views, event CRUD, tasks \\
MicroScholar & Search, papers, authors, citations, alerts \\
MicroTube    & Feed, player, comments, channels, subscriptions \\
\bottomrule
\end{tabularx}
\caption{The ten environments in \sentbench. Each is a thin clone of a real consumer product with a multi-screen UI, a typed REST API, and a catalog of synthetic data from which events are sampled.}
\label{tab:env-overview}
\end{table}

\begin{figure}[h]
    \centering
    \begin{minipage}[t]{0.28\linewidth}
        \vspace{0pt}
        \includegraphics[width=\linewidth]{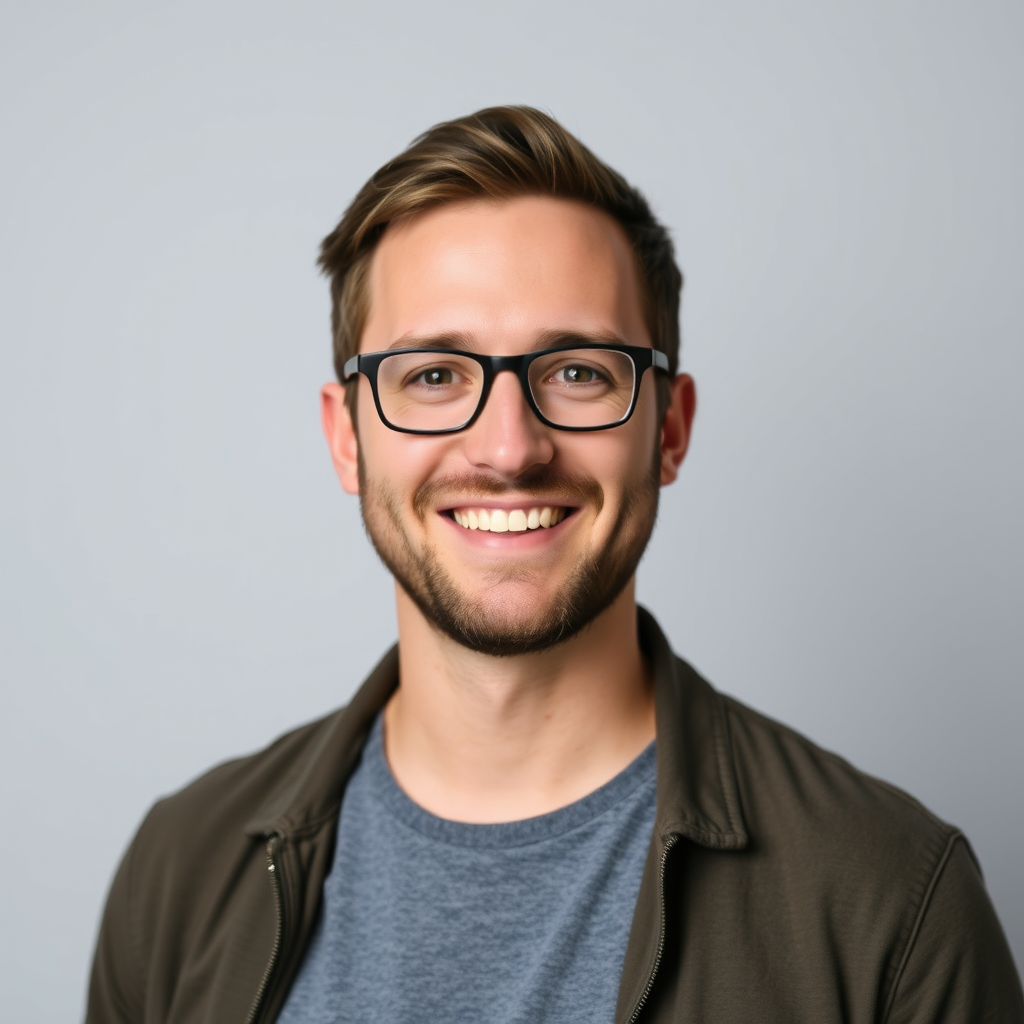}
        \vspace{0.2em}%
        \small\textbf{Chris Taylor}\\
        {\footnotesize\itshape Product Associate, 29}\\
        {\footnotesize San Francisco, CA}
    \end{minipage}\hfill
    \begin{minipage}[t]{0.68\linewidth}
        \vspace{0pt}
  {\small
  \begin{tabular}{@{}>{\ttfamily\bfseries}r @{:\hspace{6pt}} p{0.75\linewidth}@{}}
  name            & Chris Taylor \\
  username        & ctaylor \\
  email           & ctaylor@mymicromail.com \\
  age             & 29 \\
  location        & San Francisco, CA \\
  job\_title      & Product Associate \\
  bio             & A product-focused employee trying to build useful things. Juggling features, bugs, and
  way too many browser tabs. \\
  personality     & adaptable, inquisitive, team-player \\
  interests       & tech, product development, agile methodologies, rock climbing \\[6pt]
  \multicolumn{2}{@{}l}{\textbf{Per-environment profiles}} \\[2pt]
  MicroGram       & 847 followers, 312 following, 24 posts \\
  MicroScholar    & Redwood University, PhD Candidate, 1{,}569 citations \\
  MicroFy         & 156 followers, 12 playlists, 247 liked tracks \\
  MicroHub        & pinned repos, achievements, contribution history \\
  MicroDin        & TheMicroCorporate, Enterprise Software, 897 connections \\
  \end{tabular}}
    \end{minipage}
    \caption{\sentbench includes 100 synthetic user personas. Each persona consists of core demographic and biographical attributes plus per-environment subprofiles that ground the same identity coherently across all 10 simulated applications (MicroGram, MicroScholar, MicroFy,
    MicroHub, MicroDin shown). Chris Taylor, shown here, is the principal user. I.e., the user whose profile is accessed by the agent.}
    \label{fig:persona-chris}
\end{figure}

\subsection{Synthetic Data}
\label{sec:synth-data}

An email client with no emails or contacts would make for a poor simulation, as would a video-sharing website with no videos. We therefore developed a synthetic data generation pipeline to populate our environments.

The process begins by using generative models\footnote{See Appendix \ref{app:datagen} for more details about data generation.} to populate a list of 100 user personas and 201 entities, shared across all 10 environments. User personas represent fictitious people. Their profiles include names, biographical and demographic information, interests, social media presence, and other settings. User profiles can also contain multimedia assets such as avatars and profile photos. Figure~\ref{fig:persona-chris} presents the profile of Chris Taylor, the persona whose accounts are accessed by the agent in the benchmark, i.e., the agent's owner or principal user.

Conversely, entities describe companies, music bands, content creator channels, institutions, news organizations, or other outlets whose content may appear in \sentbench environments. Similar to user profiles, entity profiles can include multimedia assets, such as company logos and banner images.

Having established our universe of personas and entities, we again leverage generative models to produce environment-appropriate content. For example, with a user's profile loaded as context, we produce emails, instant messages, social media posts, playlists, and other records to populate the applications. These elements then enter a content catalog from which tasks are later constructed: some of these items pre-populate the environments, while others become scheduled events that unfold over time. In total, MicroMail ships 260 emails with realistic subjects, bodies, and attachments; MicroChat ships 200 messages across 30 conversations and 10 teams; MicroDin ships 50 posts, 35 connection requests, 25 jobs, and 12 notifications; And so forth. Additionally, catalogs can also include waypoints that allow values to vary continuously over time. This is especially important for environments like MicroHood, whose simulated stock prices are constantly changing. Pre-generating this content upfront, together with personas and entities, ensures that the environments and scenarios exhibit a degree of in-world consistency. Per-environment catalog inventories appear in Appendix~\ref{app:envs}.

\subsection{Benchmark Tasks}
\label{sec:benchmark-tasks}

With applications, personas, entities, and content in hand, we next generate the tasks that form the basis of the \sentbench benchmark. Because the usefulness of the benchmark depends directly on the quality of its tasks, we describe this step in detail. We first introduce the task dimensions that guide benchmark construction, then detail the mechanical process of generation. Finally, we describe how we validated tasks prior to inclusion in the benchmark. We summarize the tasks in Table~\ref{tab:example-tasks}, and we include a full listing of task prompts in Appendix~\ref{app:task_list}.

\subsubsection{Task Dimensions}
\label{sec:task-dimensions}

We consider two independent axes of task design: \emph{action requirement} and \emph{criterion type}, described below.

\textbf{Action Requirement}: Some meaningful tasks are achievable by passively watching for a target condition to occur, possibly followed by some minimal additional action such as contacting the user. These \emph{passive} tasks are arguably the quintessential monitoring workflow. An example of such a task is ``\emph{Watch Lithium Energy (LITH) for me and let me know when it reaches at least \$500}.'' The task is considered complete if the agent affirmatively concludes its monitoring \emph{after} the requisite event has occurred. The task fails if the agent concludes too soon (a false positive) or has not concluded by the time the simulation reaches its end. By default, all tasks are designed to be achievable within 10 minutes. Once 10 minutes is reached, agents are granted an additional 30-second grace period if they are still running, and are then terminated. As we will discuss later, a \texttt{speed\_factor} parameter (default: 1.0) can be used to scale tasks to arbitrary lengths, but the grace period is not scaled -- it remains 30 seconds regardless of whether the task is meant to take 10 minutes, 40 minutes, or something else.

In contrast, \emph{active tasks} require agents to periodically perform an action to gather new information or to respond to in-world events. An example of an active task is ``\emph{Let me know the moment Charles Davis sends me a file in any of my chats. You'll need to open each conversation to check---attachments aren't visible from the sidebar.}'' This task is considered complete if the agent affirmatively concludes its work and leaves the application database in the correct state. In this example, an attachment-bearing message from Charles Davis must exist and be marked as read. Any other outcome is considered a task failure.

Finally, we recognize that late-occurring false-positive detections may be scored as successful for passive tasks. Likewise, a malicious agent with basic knowledge of the benchmark might unconditionally conclude its monitoring just before the simulation ends, then act as if the event has occurred. For example, with 10-minute simulations, an agent might always declare that the condition has occurred after 9 minutes and 30 seconds. While such behavior would score poorly on reaction time, passive tasks would still be considered successful. To discourage such strategies and to re-balance the benchmark, we include 20 \emph{no-operation (no-op)} tasks. On the surface, these tasks resemble passive or active tasks, but we designed them such that the requisite conditions never occur. These tasks are considered successful only if the simulation ends while the agent is still working or monitoring the environment.

\begin{wraptable}{r}{0.40\textwidth}
\vspace{-0.9\baselineskip}
\centering
\begin{tabular}{@{}lcc@{}}
\toprule
 & \textbf{Absolute} & \textbf{Relative} \\
\midrule
\textbf{Passive}      & 18 & 20 \\
\textbf{Active}       & 23 & 19 \\
\textbf{No-operation} & \multicolumn{2}{c}{20} \\
\bottomrule
\end{tabular}
\caption{Task counts by action requirement and criterion type.}
\label{tab:task-counts}
\vspace{0.5\baselineskip}
\end{wraptable}

\textbf{Criterion Type}: For some agents and monitoring strategies, small changes in criterion wording can dramatically affect performance. For example, consider the task of alerting a user when a social media post crosses ``100 likes.'' In this case, a single screenshot is sufficient to determine whether the condition has been met. A memoryless agent, or an agent with aggressive context management, can achieve this task without much trouble. However, if the task is instead to alert the user when the post ``gets 100 \emph{more} likes,'' the agent must reason over an initial and an intermediate state. An agent with a long trajectory might easily forget how many likes the post started with. We refer to the first type of task as an \emph{absolute task}, and the latter as a \emph{relative task}. We include both in the benchmark.

\begin{table}[t]
\centering
\small
\renewcommand{\arraystretch}{1.25}
\begin{tabularx}{\textwidth}{@{}l X l@{}}
\toprule
\textbf{Environment} & \textbf{Task prompt} & \textbf{Categorization} \\
\midrule
MicroHub     & ``Watch the star count on this repo and let me know when it hits at least 2,000.''                                                       & passive $\cdot$ absolute \\
MicroHood    & ``Watch ChipForge Industries (CHIP) for me and let me know when it reaches at least \$520.''                                              & passive $\cdot$ absolute \\
MicroMail    & ``Keep an eye on the junk folder for me, tell me when 3 more emails end up there.''                                                       & passive $\cdot$ relative \\
MicroChat    & ``Let me know when unread messages show up in 3 more of my chats than right now. Just watch the sidebar, don't open any of the conversations, I don't want people seeing 'seen' on their messages before I've had a chance to read them.'' & passive $\cdot$ relative \\
MicroScholar & ``I've got a couple of unread alerts already, let me know when 3 more unread alerts come in.''                                          & passive $\cdot$ relative \\
MicroFy      & ``Watch the trending feed. When a song drops whose lyrics mention `subway', like it for me.''                                            & active $\cdot$ absolute \\
MicroGram    & ``Watch my DMs for a new message about a mockup, the word might not show in the preview, so you'll need to open conversations to check.'' & active $\cdot$ absolute \\
MicroLendar  & ``I'm expecting a calendar invite about a trip to Hawaii, the details should be in the event description, so you'll need to open each event to check.'' & active $\cdot$ absolute \\
MicroTube    & ``I already watched a couple videos from my subscriptions earlier. Keep watching that feed and watch 3 more new uploads for me as they appear.''            & active $\cdot$ relative \\
MicroDin     & ``Two new Python-heavy roles should come through today. Apply to both when you see them, you'll need to open the job details to check the requirements.'' & active $\cdot$ relative \\
\bottomrule
\end{tabularx}
\caption{One example task from each \sentbench environment, with categorizations for each across \textit{criterion type} and \textit{action requirement}. Appendix~\ref{app:task_list} includes the full prompts for the remaining scenarios.}
\label{tab:example-tasks}
\end{table}

\subsubsection{Task Generation}
\label{sec:task-generation}

We first define the overall composition of \sentbench, then generate tasks to fit these parameters. \sentbench consists of 10 tasks per environment, for 100 tasks overall. Of these, 20 are no-operation tasks, with 2 per environment. We divide the remaining 80 tasks approximately evenly between active tasks (42) and passive tasks (38), and separately between absolute (41) and relative tasks (39). A full breakdown is presented in table \ref{tab:task-counts}, while figure \ref{tab:example-tasks} provides an example task for each application.

For each task, we specify an environment, action requirement, criterion type, and condition time sampled uniformly from $[10,600]$ seconds. We then use an AI coding agent (Claude Code with Opus 4.6), to generate a plausible task scenario. This involves generating a prompt, sampling enough events from the data catalog to cover a 10-minute simulation, scheduling those events so that they respect the chosen condition time, and composing a SQL query used to evaluate agent success at the end of the task, e.g., to ensure the application is in the correct state. Figure~\ref{fig:sentinel_event_timeline} shows an example task.

Once generated, we subject each task to a series of deterministic automated checks to identify obvious errors. These checks resemble unit tests and ensure, for example, that passive tasks fail if concluded before the condition time but pass afterward. Likewise, we pair active tasks with their requisite actions (generated by the coding agent alongside the task), then ensure that these actions are actually necessary to pass the tests. As tasks are generated, we use rejection sampling, and regenerate any tasks that fail these basic tests. The resultant tasks are then subjected to additional manual and LLM scrutiny, as described next.

\subsubsection{Task Validation}
\label{sec:task-validation}

Even when tasks pass automated checks, they may still contain ambiguities or implementation issues. For this reason, we manually inspected every task prompt for clarity. We also attempted most tasks ourselves, in a browser, to assess their feasibility. Additionally we repeatedly ran the full benchmark with our baseline web agents, discussed later, and inspected logs for failures attributable to the tasks or environments rather than to the agents themselves. This log analysis was again performed with the help of coding agents, namely: Claude Code with Opus 4.6 and 4.7.

For example, one task asked the agent to alert the user if any instant messages mentioned failures with a site's payment-processing system. However, the scenario also included messages discussing problems on the site's checkout page. These messages were meant to serve as unrelated background events, but they created understandable ambiguity: Was the checkout page throwing an HTTP 500 error because the payment-processing system was down? We considered agent resolution of such ambiguity to be out of scope for this benchmark, and replaced the offending messages with others sampled from the content catalog.

In a separate instance, we found that the MicroScholar application did not render correctly when searches yielded zero results. We fixed this error in the environment code. 

This process of manual and AI-assisted task validation continued until successive runs surfaced few or no novel issues. While we cannot guarantee that the benchmark is now free of errors, we are confident that the tasks and environments are generally fit for purpose and provide useful signal about how agents perform on time-evolving tasks.

\section{Evaluation Protocol and Metrics}
\label{sec:evaluation}

\sentbench is designed to present agents with a natural and unopinionated interface to benchmark tasks and environments. From a benchmark user's perspective, the evaluation loop is simple: the user provides a shell command template with placeholders for the task prompt and starting webpage URL; see Figure~\ref{fig:sentinel_event_timeline}. For each task, the evaluation harness fills in these values, invokes the command, runs the simulation, and computes metrics.

This design is flexible: any web-capable agent that can be invoked from the terminal can be evaluated. The only additional requirement is that the shell process exit when the agent concludes its work. Unlike other related benchmarks or gyms~\cite{froger2026gaia2, zhou2023webarena, chezelles2024browsergym, workarena2024, xie2024osworld}, we do not require a particular agent or browser framework, such as Playwright, nor do we assume any particular observation format or modality, such as screenshots, accessibility trees, or the document object model. However, enabling this level of flexibility places additional requirements on the evaluation harness and metrics, which we describe in the remainder of this section.

\subsection{Simulation Life Cycle}
\label{sec:lifecycle}

As noted earlier, each \sentbench environment includes a FastAPI backend server that serves two purposes: First, it powers the web applications, for example by serving search results and populating social feeds. Second, it works with the evaluation harness to move each scenario through its life cycle. %
Figure~\ref{fig:sentinel_api_diagram} illustrates these state transitions, and we provide additional details below:

\begin{figure}[t] 
    \centering
    \includegraphics[width=\columnwidth]{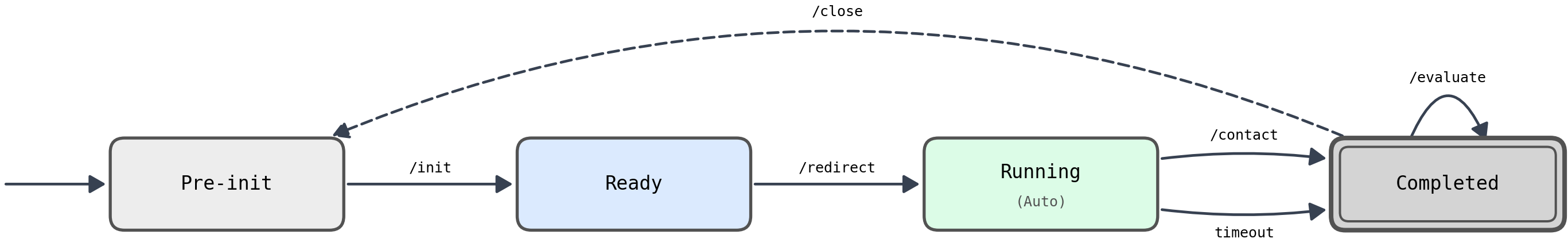}
  \caption{The simulation life cycle for \sentbench environments. The evaluation harness interacts with the server to transition between life cycle states.}
    \label{fig:sentinel_api_diagram} 
\end{figure}

\textbf{Pre-initialization}: The \sentbench environment server begins in the pre-initialization state, in which all environment databases are in their initial conditions and no events are queued for playback. The evaluation harness loads a scenario from disk, then calls the server's \texttt{/init} endpoint (e.g., \texttt{http://localhost:8000/init}) and provides the parameters of the scenario. Parameters include: updates to the initial database state, the agent's starting page, a list of events to play back, a \texttt{speed\_factor} parameter to control the playback speed, and SQL query to evaluate task success. The server then transitions to the \emph{Ready} state.

\textbf{Ready}: Upon receiving confirmation that the server has entered the Ready state, the evaluation harness invokes the agent with the scenario task prompt. The prompt is augmented with instructions to facilitate evaluation. Specifically, agents are instructed to complete a web form at the end of the task to notify the principal user, thus signaling task completion. 

We recognize that agents can take time to start up, especially when browsers or sandboxes are involved. To avoid counting this startup time against the agent, the evaluation harness initially directs the agent to a \texttt{/redirect} endpoint (e.g., \texttt{http://localhost:8000/redirect}) as the starting webpage URL. When the agent accesses this URL, the server responds with an HTTP 301 redirect to the true task starting page, previously loaded during pre-initialization. This triggers two concurrent actions: the agent's browser follows the redirect, and the server enters the \emph{Running} state.

\textbf{Running}: In the Running state, the server plays back all scheduled events at their corresponding times. These times are computed by dividing each event's default time by an optional \texttt{speed\_factor} parameter. For example, if an event was originally meant to occur after $30$ seconds, then a \texttt{speed\_factor} of $2.0$ schedules it to occur $15$ seconds into the simulation (i.e., the simulation runs twice as fast). The \texttt{speed\_factor} defaults to $1.0$, giving a 10 minute task, but can be scaled to any desired interval.

When events occur, they update the application database and are reflected in the interface. If the simulation reaches its natural end, the server initiates a fixed 30-second grace period (i.e., independent of the \texttt{speed\_factor}) before transitioning to the \emph{Completed} state.

The server also transitions to Completed if the agent accesses the \texttt{/contact} form (e.g., \path{http://localhost:8000/contact}), signaling that the agent believes the task is done. This cancels all remaining scheduled events, even if the transition occurs well before the scenario end time.

\textbf{Completed}: Upon detecting that the server has entered the Completed state, the evaluation harness accesses the \texttt{/evaluate} endpoint (e.g., \texttt{http://localhost:8000/evaluate}). This endpoint executes the previously saved SQL query against the database to determine whether the task was successful. It also returns timing metadata, such as the number of seconds the simulation spent in the Running state. These values are recorded and used to compute benchmark metrics after the final task.

Finally, the evaluation harness accesses a \texttt{/close} endpoint (e.g., \texttt{http://localhost:8000/close}), which resets the database states and transitions the server back to the pre-initialization state, ready for the next task.

\subsection{Evaluation Metrics}
\label{sec:evaluation_metrics}

When \sentbench is run, it creates a \texttt{results} folder to store execution artifacts. Within this folder, each task receives its own subdirectory, containing the agent's terminal output and the results data returned by the server's \texttt{/evaluate} endpoint at the end of the simulation. Because the benchmark is permissive about agents, it cannot assume privileged knowledge of token usage or cost. Agents may therefore produce an optional \texttt{costs.json} sidecar file declaring the number of input tokens, output tokens, and tool calls used during the task. When present, this file is used to compute token-related metrics. We describe these and other metrics below.

\textbf{Success.} A task's success criterion depends on its action type, as enumerated earlier. An agent succeeds at a passive task if it accesses the \texttt{/contact} form at any point after the target event has occurred. Similarly, an active task succeeds if the database is in the correct state when evaluated, as determined by the task-specific SQL query. Finally, no-operation tasks succeed if the simulation ends without the agent ever opening the \texttt{/contact} form. 

\textbf{Task Completion Time and Reaction Time.} We define a task's completion time as the total number of seconds the simulation spends in the Running state, prior to the agent accessing the \texttt{/contact} endpoint or timing out. Likewise, a scenario's target time is the number of seconds into a simulation at which the target condition (Figure~\ref{fig:sentinel_event_timeline}, \texttt{condition\_at}) is met. Reaction time is then the difference between the completion time and the target time. For example, a reaction time of 30 seconds indicates that the agent accessed the contact form 30 seconds after the target condition was first met. Reaction time cannot be meaningfully computed for no-operation tasks, since no event ever meets the requisite task condition. %

\textbf{Input Token Utilization, Output Token Utilization, and Monetary Costs.}
Another important category of metrics is inference cost. When agents self-report their token utilization, we track the cost of each task in terms of input tokens, output tokens (including reasoning tokens), and the total monetary cost charged by the inference provider. Monetary cost is an especially useful measure because it meaningfully combines input and output tokens into a single value, accounting for the fact that output tokens are often charged at a higher rate than input tokens. It also allows direct comparisons across models (e.g., a weaker model may use more tokens, but is cheaper overall).

Having described task lifecycle, and metrics, we now present an initial set of evaluations.

\section{Baseline Evaluations}
\label{sec:baseline_evaluations}

We used \sentbench to compare three models, and two agent configurations. The purpose of these evaluations is twofold: First, we demonstrate that \sentbench can meaningfully distinguish between models and agents. Second, the evaluations establish baselines from which to compare future benchmarks scores. We detail these experiment conditions below, then present results, and provide some additional error analysis of common agent failures, as detected by the benchmark.

\subsection{Experiment Conditions}
For our baselines, we pair three multimodal models: GPT-5.4 with `low' reasoning, Qwen 3.5:9B, and GPT-4o, with a web browsing agent adapted from our work on Magentic-UI  
\cite{mozannar2025magenticui}. Here, the agent follows a simple tool-calling loop: At each iteration, the model receives a screenshot of the current browser state along with a brief description, then selects from a set of browser tools such as \emph{click}, \emph{type}, \emph{scroll}, and \emph{visit URL}. The loop continues until the model stops making tool calls or invokes a \emph{terminate} tool.

For this evaluation, we compare two variations of the agent toolset. In the baseline condition, the toolset includes a \texttt{sleep(time)} tool. When invoked, this tool unconditionally blocks the execution thread for the specified number of seconds before returning control to the agent loop.

In the second condition, agents are equipped with a purpose-built \texttt{wait\_for(condition, timeout)} tool. This tool takes a natural-language condition and a timeout specifying the maximum amount of time it may block before returning. For example, \texttt{wait\_for(``a new email to arrive'', 300)} blocks until a new email is detected or 300 seconds have elapsed.

The implementation of \texttt{wait\_for} is intentionally simple: When invoked, the tool first captures a textual snapshot of the page to use as a baseline. It then enters a fast loop, capturing a new textual snapshot once per second. On each iteration, the Python \texttt{difflib} library computes a unified diff between the baseline and current page state. The resulting diff consists of blocks, each representing a contiguous section of the page that has changed. Blocks seen  in previous iterations are removed, on the assumption that these changes have already been evaluated. The remaining blocks are passed to the LLM with a prompt asking whether the new changes satisfy the target condition. The loop breaks if the answer is `yes', and continues if the answer is `no'. The \texttt{wait\_for} tool also reloads the webpage periodically to capture changes on static pages, and enforces moderate rate limits to prevent thrashing on fast-changing websites. A full pseudo code listing is provided in Appendix~\ref{app:wait_for}

\subsection{Results}

Across all six conditions (three models, each paired with either \texttt{wait\_for} or \texttt{sleep}), we report three complementary measures of performance: how often agents complete tasks (\emph{success rate}), how much they spend doing so (\emph{per-task cost}), and how quickly they respond once a condition is met (\emph{reaction time}). Reporting these together is deliberate: an agent can look strong on success rate alone while being far too expensive or too slow to be practical, and it is precisely this tension between responsiveness and cost that \sentbench is built to expose. We examine each measure in turn below, then show in Section~\ref{sec:speed_factor} how the gaps between conditions widen as tasks are stretched to require longer waits.

\subsubsection{Success Rate}
With three models and two agent configurations, we evaluate a total of six conditions. We first consider the success rates of each condition, as presented in Table~\ref{tab:success_rate}. As expected, GPT-5.4, in the `low' reasoning setting, performs consistently and substantially better (0.75 with \texttt{wait\_for} and 0.68 with \texttt{sleep}) than the much smaller Qwen 3.5:9B model (0.48 and 0.49) and the much older GPT-4o model (0.48 and 0.46).

Also as expected, we find that passive tasks have higher success rates than active tasks in all conditions, suggesting that they are generally easier to accomplish. We also find that both Qwen 3.5:9B and GPT-4o perform substantially better when tasks use absolute prompt phrasing. Somewhat unexpectedly, GPT-5.4 performs slightly, but consistently, better on tasks that use relative phrasing.

Finally, the choice between \texttt{sleep} and \texttt{wait\_for} matters little for Qwen 3.5:9B and GPT-4o. However, when GPT-5.4 uses the \texttt{sleep} tool, it fails an additional five no-operation tasks, accounting for most of the 7-point performance difference between these conditions. Inspection of the logs shows that when GPT-5.4 uses the \texttt{sleep} tool, it occasionally concludes monitoring early, even while recognizing that the condition has not been met. This is evident from the agent trajectories, with GPT-5.4 concluding tasks with messages such as ``\emph{I checked the chats and did not find any conversation where Diana Miller @mentioned you.}''

\begin{table}[t]
\centering
\small
\begin{tabularx}{\textwidth}{@{}l *{6}{>{\centering\arraybackslash}X}@{}}
\toprule
& \textbf{Overall} & \multicolumn{3}{c}{\textbf{Action Requirement}} & \multicolumn{2}{c}{\textbf{Criterion Type}} \\
\cmidrule(lr){3-5}
\cmidrule(l){6-7}
\textbf{Condition} & \textbf{Success Rate} & \textbf{No-op} & \textbf{Passive} & \textbf{Active} & \textbf{Absolute} & \textbf{Relative} \\
\midrule
GPT-5.4, \texttt{wait\_for}     & 0.75 & 0.95 & 0.92 & 0.50 & 0.64 & 0.77 \\
GPT-5.4, \texttt{sleep}         & 0.68 & 0.70 & 0.76 & 0.60 & 0.64 & 0.72 \\
GPT-4o, \texttt{wait\_for}      & 0.48 & 0.95 & 0.63 & 0.12 & 0.50 & 0.23\\
GPT-4o, \texttt{sleep}          & 0.46 & 1.00 & 0.53 & 0.14 & 0.33 & 0.33 \\
Qwen 3.5:9b, \texttt{wait\_for} & 0.48 & 0.95 & 0.50 & 0.24 & 0.45 & 0.28 \\
Qwen 3.5:9b, \texttt{sleep}     & 0.49 & 0.95 & 0.39 & 0.36 & 0.48 & 0.28 \\
\bottomrule
\end{tabularx}
\caption{Success rates for three models and two agent configurations (six conditions total). As expected, GPT-5.4 performs better overall than GPT-4o and Qwen 3.5:9B. Likewise, agents configured to use \texttt{wait\_for} perform about as well as, or better than, agents configured to use \texttt{sleep}. We also observe that no-operation tasks are generally easier than passive tasks, and that active tasks are harder still. The one exception is GPT-5.4 with \texttt{sleep}, which performs unexpectedly poorly on no-operation tasks.}
\label{tab:success_rate}
\end{table}

\subsubsection{Per-task Cost}
\label{sec:tool_cost}

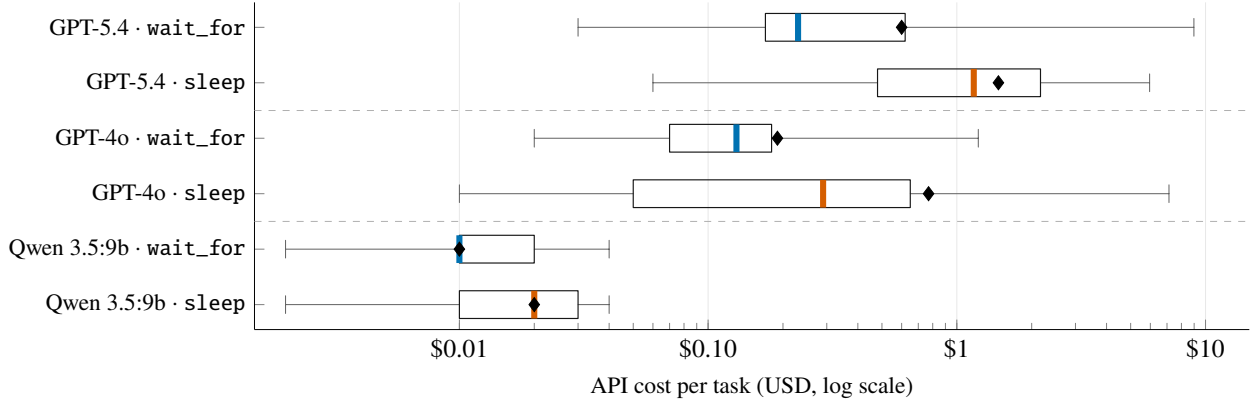
\begin{figure}[t]
\centering
\resizebox{\linewidth}{!}{%
\begin{tikzpicture}
\begin{axis}[
    width=\linewidth, height=6.5cm,
    boxplot/draw direction=x,
    xmode=log, xmin=0.0015, xmax=15,
    xtick={0.01,0.1,1,10},
    xticklabels={\$0.01,\$0.10,\$1,\$10},
    xlabel={API cost per task (USD, log scale)},
    xlabel style={font=\small},
    xmajorgrids=true,
    grid style={gray!18, line width=0.4pt},
    axis x line*=bottom, axis y line*=left,
    ytick={1,2,3,4,5,6},
    yticklabels={
        {Qwen 3.5:9b $\cdot$ \texttt{sleep}},
        {Qwen 3.5:9b $\cdot$ \texttt{wait\_for}},
        {GPT-4o $\cdot$ \texttt{sleep}},
        {GPT-4o $\cdot$ \texttt{wait\_for}},
        {GPT-5.4 $\cdot$ \texttt{sleep}},
        {GPT-5.4 $\cdot$ \texttt{wait\_for}},
    },
    yticklabel style={font=\small},
    ymin=0.55, ymax=6.45,
    enlarge y limits=false, clip=false,
]
\addplot[draw=none, mark=none, boxplot prepared={draw position=6,
    lower whisker=0.03, lower quartile=0.17, median=0.23,
    upper quartile=0.62, upper whisker=8.98, average=0.60,
    box extend=0.5, whisker extend=0.32,
    every box/.style={draw=cink, fill=none, line width=0.5pt, line cap=round, line join=round},
    every median/.style={draw=cblue, line width=2.6pt},
    every whisker/.style={draw=cmuted, line width=0.35pt, line cap=round},
    every average/.style={mark=diamond*, mark size=3pt,
        mark options={fill=white, draw=cink, line width=0.5pt}}}] coordinates {};
\addplot[draw=none, mark=none, boxplot prepared={draw position=5,
    lower whisker=0.06, lower quartile=0.48, median=1.17,
    upper quartile=2.17, upper whisker=5.96, average=1.47,
    box extend=0.5, whisker extend=0.32,
    every box/.style={draw=cink, fill=none, line width=0.5pt, line cap=round, line join=round},
    every median/.style={draw=corange, line width=2.6pt},
    every whisker/.style={draw=cmuted, line width=0.35pt, line cap=round},
    every average/.style={mark=diamond*, mark size=3pt,
        mark options={fill=white, draw=cink, line width=0.5pt}}}] coordinates {};
\addplot[draw=none, mark=none, boxplot prepared={draw position=4,
    lower whisker=0.02, lower quartile=0.07, median=0.13,
    upper quartile=0.18, upper whisker=1.22, average=0.19,
    box extend=0.5, whisker extend=0.32,
    every box/.style={draw=cink, fill=none, line width=0.5pt, line cap=round, line join=round},
    every median/.style={draw=cblue, line width=2.6pt},
    every whisker/.style={draw=cmuted, line width=0.35pt, line cap=round},
    every average/.style={mark=diamond*, mark size=3pt,
        mark options={fill=white, draw=cink, line width=0.5pt}}}] coordinates {};
\addplot[draw=none, mark=none, boxplot prepared={draw position=3,
    lower whisker=0.01, lower quartile=0.05, median=0.29,
    upper quartile=0.65, upper whisker=7.13, average=0.77,
    box extend=0.5, whisker extend=0.32,
    every box/.style={draw=cink, fill=none, line width=0.5pt, line cap=round, line join=round},
    every median/.style={draw=corange, line width=2.6pt},
    every whisker/.style={draw=cmuted, line width=0.35pt, line cap=round},
    every average/.style={mark=diamond*, mark size=3pt,
        mark options={fill=white, draw=cink, line width=0.5pt}}}] coordinates {};
\addplot[draw=none, mark=none, boxplot prepared={draw position=2,
    lower whisker=0.002, lower quartile=0.01, median=0.01,
    upper quartile=0.02, upper whisker=0.04, average=0.01,
    box extend=0.5, whisker extend=0.32,
    every box/.style={draw=cink, fill=none, line width=0.5pt, line cap=round, line join=round},
    every median/.style={draw=cblue, line width=2.6pt},
    every whisker/.style={draw=cmuted, line width=0.35pt, line cap=round},
    every average/.style={mark=diamond*, mark size=3pt,
        mark options={fill=white, draw=cink, line width=0.5pt}}}] coordinates {};
\addplot[draw=none, mark=none, boxplot prepared={draw position=1,
    lower whisker=0.002, lower quartile=0.01, median=0.02,
    upper quartile=0.03, upper whisker=0.04, average=0.02,
    box extend=0.5, whisker extend=0.32,
    every box/.style={draw=cink, fill=none, line width=0.5pt, line cap=round, line join=round},
    every median/.style={draw=corange, line width=2.6pt},
    every whisker/.style={draw=cmuted, line width=0.35pt, line cap=round},
    every average/.style={mark=diamond*, mark size=3pt,
        mark options={fill=white, draw=cink, line width=0.5pt}}}] coordinates {};
\draw[cmuted!50, line width=0.3pt, dashed] (axis cs:0.0015,4.5) -- (axis cs:15,4.5);
\draw[cmuted!50, line width=0.3pt, dashed] (axis cs:0.0015,2.5) -- (axis cs:15,2.5);
\end{axis}
\end{tikzpicture}}%
\caption{Per-task API cost in USD for each model and tool configuration. Box outlines show the interquartile range, the colored bar inside each box marks the median (blue for \texttt{wait\_for}, orange for \texttt{sleep}), whiskers extend to the observed minimum and maximum, and the diamond marks the mean. Dashed horizontal lines separate model groups. The horizontal axis is logarithmic; Qwen 3.5:9B is roughly two orders of magnitude cheaper than the GPT models, and within every model \texttt{sleep} is consistently more expensive than \texttt{wait\_for}. Means landing outside Q3 (GPT-5.4 \texttt{wait\_for}, GPT-4o \texttt{sleep}) reflect right-skewed distributions with high-cost outliers visible at the upper whisker.}
\label{fig:task_cost}
\end{figure}

The differences between models, and between \texttt{wait\_for} and \texttt{sleep}, become even clearer when we examine per-task cost. Figure~\ref{fig:task_cost} presents these results, computed from API prices posted on May 15, 2026. For GPT models, we use prices from \texttt{https://developers.openai.com}; for Qwen 3.5:9B, we use prices from \texttt{https://openrouter.ai}.

Across all conditions, \texttt{sleep} is substantially more expensive than \texttt{wait\_for}. With GPT-5.4, the median task cost is $5.1\times$ higher with \texttt{sleep} than with \texttt{wait\_for} ($\$1.17$ vs. $\$0.23$). Similarly, \texttt{sleep} is about $2.2\times$ and $2.0\times$ as costly as \texttt{wait\_for} for GPT-4o and Qwen 3.5:9B, respectively ($\$0.29$ vs. $\$0.13$; and $\$0.02$ vs. $\$0.01$). Inspecting the logs, we find that agents configured with \texttt{sleep} either sleep for very brief intervals, such as 5--10 seconds at a time, or do not sleep at all. This leads to longer agent trajectories, as reflected in Table~\ref{tab:tool_calls}. For example, GPT-5.4 makes a median of 6 tool calls with \texttt{wait\_for}, compared with 19.5 with \texttt{sleep}. Given that \texttt{wait\_for} performs as well as or better than \texttt{sleep} on task completion, these cost differences suggest little downside to using \texttt{wait\_for}. More importantly, these findings highlight the importance of including tokens costs as a metric in this benchmark.

\begin{table}[t]
\centering
\small
\begin{tabularx}{\textwidth}{@{}l *{6}{>{\centering\arraybackslash}X}@{}}
\toprule
& \multicolumn{6}{c}{\textbf{Number of Tool Calls}} \\
\cmidrule(l){2-7}
\textbf{Condition} & \textbf{Mean} & \textbf{Median} & \textbf{Q1} & \textbf{Q3} & \textbf{Min} & \textbf{Max} \\
\midrule
GPT-5.4, \texttt{wait\_for}     & 9.6 & 6 & 5 & 12 & 1 & 42 \\
GPT-5.4, \texttt{sleep}         & 20.9 & 19.5 & 12.25 & 28.75 & 3 & 47 \\
GPT-4o, \texttt{wait\_for}      & 4.9 & 5 & 2 & 6 & 1 & 21 \\
GPT-4o, \texttt{sleep}          & 11.9 & 9.5 & 3 & 14 & 0 & 50 \\
Qwen 3.5:9b, \texttt{wait\_for} & 13.8 & 11.5 & 6 & 21 & 1 & 32 \\
Qwen 3.5:9b, \texttt{sleep}     & 17.0 & 16 & 8.25 & 26 & 1 & 33 \\
\bottomrule
\end{tabularx}
\caption{
Descriptive statistics of the number of tools calls made in each of the six conditions. \texttt{sleep} uses more tool calls, and leads to longer trajectories, in both the mean and median cases.}
\label{tab:tool_calls}
\end{table}

\subsubsection{Reaction Time}
\label{sec:reaction_time}
In the previous section, we reported that \texttt{wait\_for} is substantially cheaper than \texttt{sleep}, but these savings could conceivably come at the cost of slower agent reaction times. To address this concern, Table~\ref{tab:reaction_time} reports the mean and median agent reaction times for successful tasks. Here, we exclude no-operation tasks and failed tasks, because reaction time is poorly defined under these conditions.

Overall, we find a mixed signal. For GPT-5.4, the median reaction time is 9.4 seconds slower when using \texttt{wait\_for} than when using \texttt{sleep} (51.7 seconds vs. 42.3 seconds). However, \texttt{wait\_for} is nearly twice as fast for GPT-4o and Qwen 3.5:9B (22.8 seconds vs. 48.6 seconds, and 60.1 seconds vs. 123.8 seconds, respectively).

Here, we note that direct comparisons across models are less meaningful because inference times depend on the characteristics of their respective endpoints. For example, Qwen 3.5:9B's reaction times are longest because we hosted this model locally and achieved lower token throughput than the production GPT endpoints.

\begin{table}[t]
\centering
\small
\begin{tabularx}{\textwidth}{@{}l *{6}{>{\centering\arraybackslash}X}@{}}
\toprule
& \multicolumn{2}{c}{\textbf{Reaction Time in Seconds}} \\
\cmidrule(l){2-3}
\textbf{Condition} & \textbf{Mean} & \textbf{Median} \\
\midrule
GPT-5.4, \texttt{wait\_for}     & 81.4~s & 51.7~s  \\
GPT-5.4, \texttt{sleep}         & 73.0~s & 42.3~s \\
GPT-4o, \texttt{wait\_for}      & 35.1~s & 22.8~s \\
GPT-4o, \texttt{sleep}          & 59.7~s & 48.6~s \\
Qwen 3.5:9b, \texttt{wait\_for} & 100.6~s & 60.1~s  \\
Qwen 3.5:9b, \texttt{sleep}     & 140.9~s & 123.8~s \\
\bottomrule
\end{tabularx}
\caption{Mean and median reaction times for successful tasks across all six conditions. Unsuccessful tasks and no-operation tasks are excluded because reaction time is not well-defined in these cases. Comparisons within each model family are more informative than comparisons across model families, since each inference endpoint may have different token throughput or may be subject to different load. With GPT-5.4, \texttt{wait\_for} responds 8.4 seconds more slowly on average than \texttt{sleep}. In the other conditions, \texttt{wait\_for} is the faster of the two tools.}
\label{tab:reaction_time}
\end{table}

\subsection{Impact of Task Duration (speed factor)}
\label{sec:speed_factor}
Finally, we are interested in how the \texttt{speed\_factor} impacts baseline performance. Recall that, by default, \sentbench tasks are designed to be completable within 10 minutes. If an agent has not completed a task within this time, the process is terminated and the results are evaluated. The choice to use a 10-minute period was a compromise: it is long enough to differentiate agents on monitoring tasks, yet short enough that the benchmark can be run in a reasonable amount of time.

However, trends observed in earlier experiments suggest that differences between agents become even more stark as tasks grow longer and agents are required to wait more. To evaluate this, we lower the benchmark \texttt{speed\_factor} parameter from $1.0$ to $0.25$, so that tasks may require as long as 40 minutes to complete. We then re-evaluate the best-performing model, GPT-5.4, with both \texttt{wait\_for} and \texttt{sleep}. The results of this experiment are summarized in Table~\ref{tab:speed_025}.

\begin{wraptable}{r}{0.46\textwidth}
\vspace{-0.9\baselineskip}
\centering
\small
\begin{tabular}{@{}l c c@{}}
\toprule
\textbf{Measurement} & \textbf{\texttt{wait\_for}} & \textbf{\texttt{sleep}} \\
\midrule
Success Rate & 0.69 & 0.56  \\
Median API Cost (USD) & $\$0.48$ & $\$4.65$  \\
Median Reaction Time & 54.8\,s & 38.9\,s  \\
\bottomrule
\end{tabular}
\caption{GPT-5.4 with \texttt{wait\_for} vs.\ \texttt{sleep} at \texttt{speed\_factor}\,$=0.25$ (tasks stretched to as long as 40 minutes). In this setting, the agent solves 13 more tasks when using \texttt{wait\_for} than when using \texttt{sleep}, despite the former being $9.7\times$ cheaper.}
\label{tab:speed_025}
\vspace{0.5\baselineskip}
\end{wraptable}

\begin{figure}[t]
\centering

\begin{subfigure}[t]{0.49\textwidth}
    \centering
    \includegraphics[width=\linewidth]{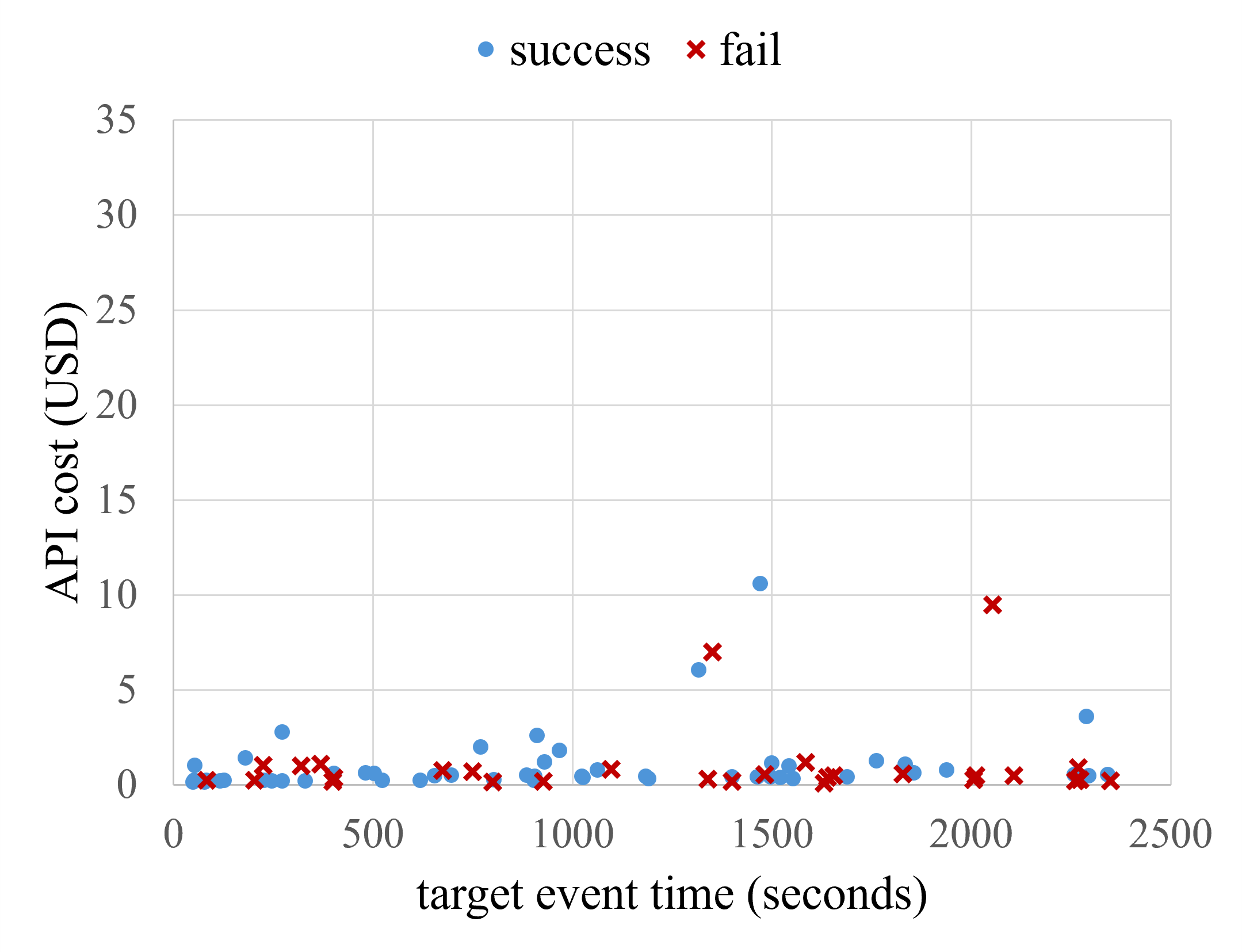}
    \caption{API cost vs. target event time (GPT-5.4; \texttt{wait\_for})}
    \label{fig:gpt55-wait-for-cost}
\end{subfigure}
\hfill
\begin{subfigure}[t]{0.49\textwidth}
    \centering
    \includegraphics[width=\linewidth]{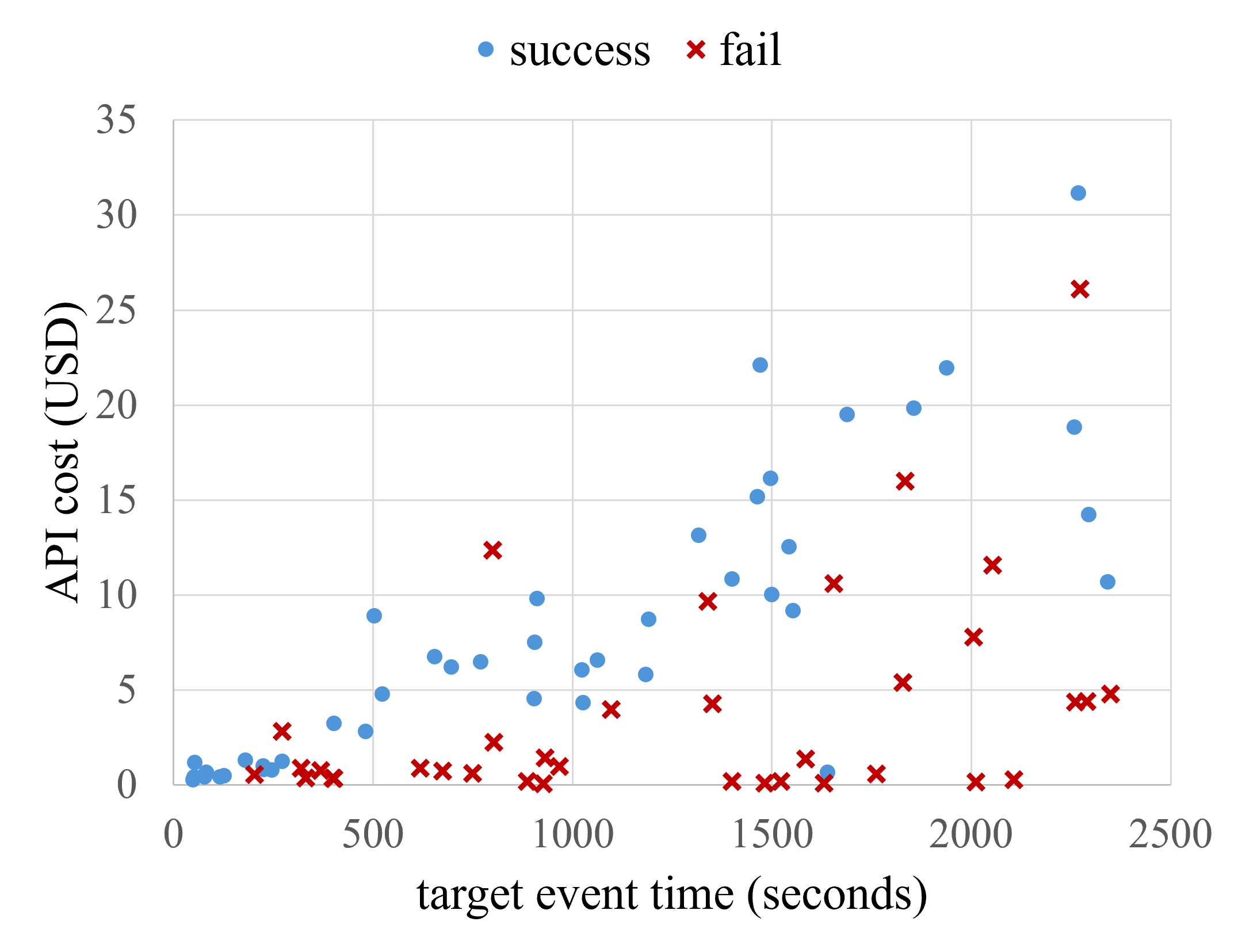}
    \caption{API cost vs. target event time (GPT-5.4; \texttt{sleep})}
    \label{fig:gpt55-sleep-cost}
\end{subfigure}

\caption{Per-task API cost as a function of target event time for GPT-5.4 under the \texttt{wait\_for} and \texttt{sleep} tool configurations. Successful tasks appear as blue dots. Failed tasks appear as red `X's. Here, tasks are scaled to take up to 40 minutes to complete (2400 seconds). When using \texttt{wait\_for}, costs remain relatively stable and low, with some outliers as high as $\$10.59$. When using \texttt{sleep}, costs trend upward with time, especially for successful tasks, with costs as high as $\$31.15$.}
\label{fig:gpt55-cost-vs-target-time}
\end{figure}

\begin{figure}[t]
\centering

\begin{subfigure}[t]{0.49\textwidth}
    \centering
    \includegraphics[width=\linewidth]{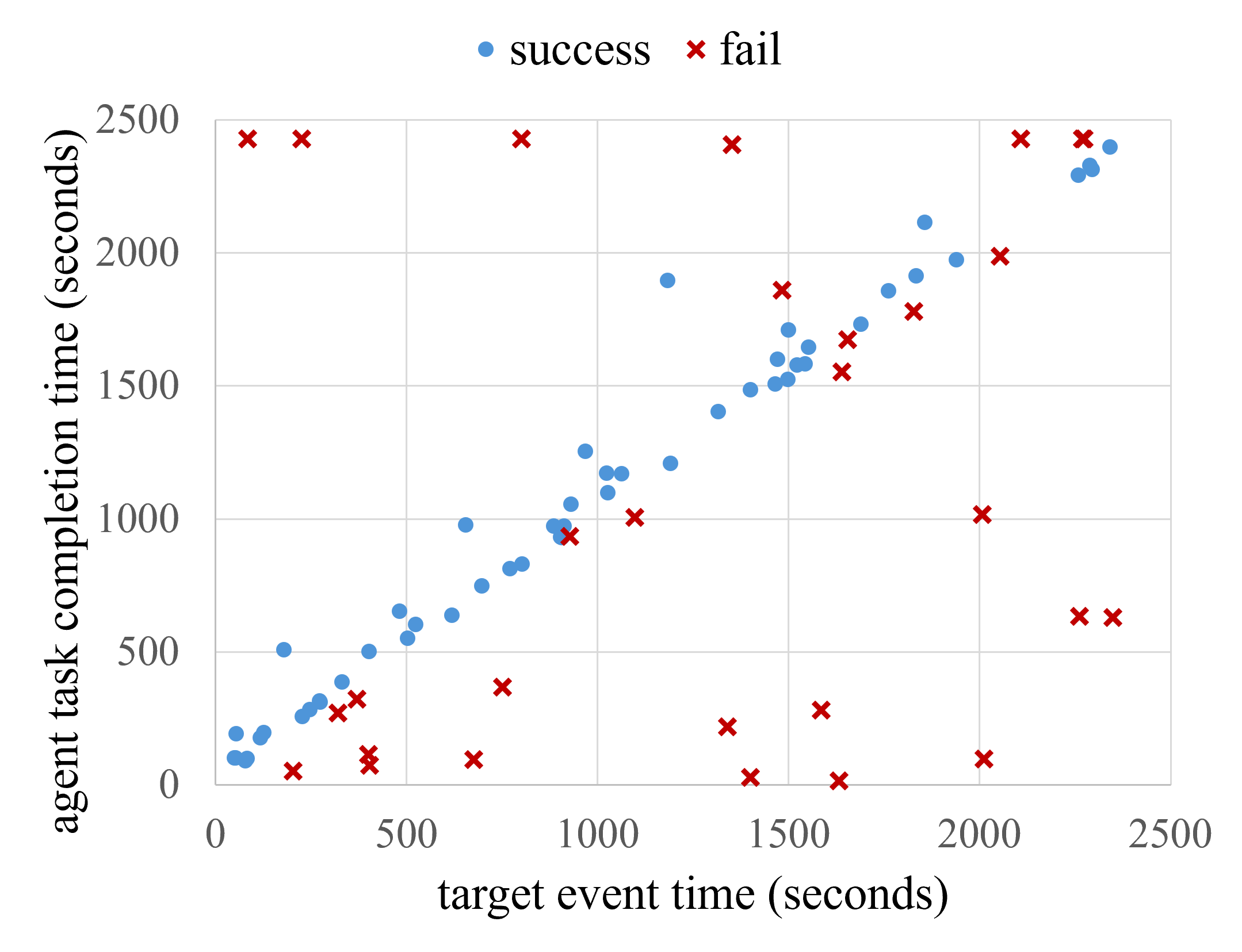}
    \caption{Task completion time vs. target event time (GPT-5.4; \texttt{wait\_for})}
    \label{fig:gpt55-wait-for-task-time}
\end{subfigure}
\hfill
\begin{subfigure}[t]{0.49\textwidth}
    \centering
    \includegraphics[width=\linewidth]{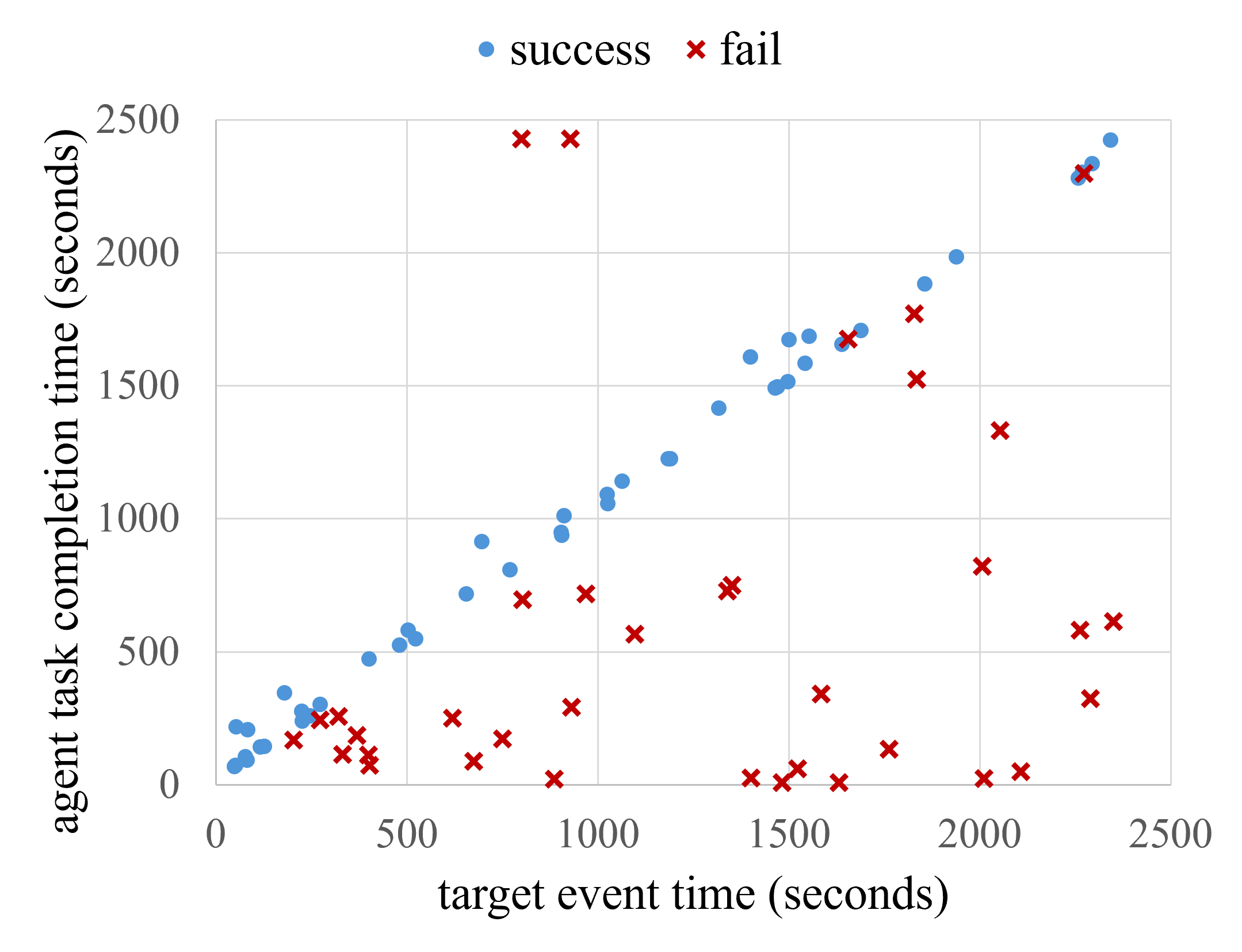}
    \caption{Task completion time vs. target event time (GPT-5.4; \texttt{sleep})}
    \label{fig:gpt55-sleep-task-time}
\end{subfigure}

\caption{Per-task completion time as a function of target event time for GPT-5.4 under the \texttt{wait\_for} and \texttt{sleep} tool configurations. Again, successful tasks appear as blue dots. Failed tasks appear as red `X's. The closer a successful task falls to the diagonal $y=x$, the better the agent's reaction time to that event. Points below this line are failures resulting from premature termination of the task. Agents are terminated after 2430 seconds, if not already concluded. Points appearing at this $y$-value indicate failed event detections. In both conditions, the diagonal is prominent, indicating that reaction times are often fast when events are detected. However, \texttt{sleep} is prone to terminating too early. Conversely, \texttt{wait\_for} performs better overall, but is slightly more prone to missing events and thus terminating late.}
\label{fig:gpt55-task-time-vs-target-time}
\end{figure}

When tasks require 40 minutes of waiting, the median per-task API cost of \texttt{sleep} ($\$4.65$) is $9.7\times$ greater than \texttt{wait\_for} ($\$0.48$), and 13 fewer tasks are completed successfully overall (56 out of 100, versus 69 out of 100). This story grows even more clear in Figures~\ref{fig:gpt55-cost-vs-target-time} and~\ref{fig:gpt55-task-time-vs-target-time}, which plot API cost, and task completion time, versus target condition time, respectively. Specifically, Figure~\ref{fig:gpt55-cost-vs-target-time}b shows that the cost for a successful task grows steadily when using \texttt{sleep}, but remains relatively constant when using \texttt{wait\_for} (Figure~\ref{fig:gpt55-cost-vs-target-time}a).

Similarly, Figure~\ref{fig:gpt55-task-time-vs-target-time} plots the agent's task completion time against the time at which the necessary event occurred, i.e., the target time. The closer a dot falls to the line $y=x$, the better the agent's reaction time. From this graph, we can clearly observe that most failures in the \texttt{sleep} condition (Figure~\ref{fig:gpt55-task-time-vs-target-time}b) correspond to negative reaction time, meaning the agent gave up on the task too early.

From this we conclude that, when when agents use the same model, their tooling can dramatically impact performance and costs. \sentbench clearly exposes these differences.

\section{Discussion and Limitations}
\label{sec:discussion}

In this paper, we introduced \sentbench and demonstrated that it can measure meaningful differences between models or agents. Like any benchmark, it makes deliberate tradeoffs between realism and controlled evaluation, in our case around task duration, and scenario diversity. In this section, we outline these limitations, highlight opportunities for improvement, and discuss future work.

A first limitation of \sentbench is that event timing is artificial. Tasks are built around a 10-minute window,  target events are scheduled randomly, and other event timings are selected by an LLM during task generation. While this is sufficient for creating dynamic environments and challenging agents with monitoring tasks, future versions may require realistic timing distributions (e.g., achievable by sampling them from real online systems). As an example, a future model might anticipate how often different page elements update, and  allocate resources accordingly. Artificial timings would interfere with such strategies.

Relatedly, \sentbench environments are lightweight facsimiles of popular websites. Prolonged exploration may expose the edges of these environments. We designed tasks to focus on areas of the environments with good feature coverage, but researchers authoring new tasks may need to extend the environments with new features. We also cannot guarantee that the environments are error-free. Because our debugging process was greedy (driven by baseline runs and manual inspection), remaining errors are most likely to occur in trajectories that diverge substantially from the paths we tested. 

Several task dimensions also remain underexplored. Although \sentbench includes both absolute and relative task phrasings, most success criteria are objective, involving specific numeric targets, entities, or topics (e.g., ``when the repo hits 2,000 stars''). Future tasks could introduce more subjective criteria (e.g., ``when any urgent bug reports come in''), requiring agents to make more subjective judgments about the relevance of events to specific criteria. Similarly, most current tasks monitor for persistent conditions, i.e., those that are unlikely to revert once they become true. For instance, once an email or instant message has arrived, it typically remains visible in the conversation history unless later moved or deleted by the user. In the future we may consider tasks that monitor for ephemeral conditions. Here, the target condition may hold only briefly, and a missed opportunity may make the task impossible to complete (e.g., ``buy the CHIP stock when it dips below \$500'').

Finally, \sentbench may also be useful for training agents, not just for evaluating them. However, two pieces are needed to support this use case. First, task generation must scale. This likely means eliminating any remaining manual verification and debugging steps. Second, roll outs will be to be compressed or accelerated, so agents do not actually spend 10 (or 40) minutes on each task. GAIA2 \cite{froger2026gaia2} offers one possible time acceleration strategy, advancing simulation time to the next event whenever the agent sleeps. In \sentbench, this is more complicated because agents often control full web browsers, and these clients often rely on time to verify certificates, pace animations or page updates, or render relative timestamps through JavaScript (e.g.,  ``posted 5 minutes ago''.) Compressing simulation time would therefore require all clients and components to synchronize against a common clock.

In summary, \sentbench provides a controlled foundation for studying time-evolving monitoring tasks, but it is only a starting point. Future work should improve event timing realism, broaden task criteria, support ephemeral events, and make the environments more scalable for both evaluation and training. These extensions would move \sentbench from a focused benchmark toward a broader testbed for long-running agent behavior.

\section{Related Work}
\label{sec:related-work}
Most benchmarks for agentic systems assume reactive environments, where state changes occur only in response to the agent's actions. We review four categories of such benchmarks and identify the measurement gaps that \sentbench addresses. The closest related work is ARE \citep{froger2025scaling}, which departs from the reactive paradigm with an event-driven simulation environment where the world evolves independently of the agent. We situate \sentbench relative to ARE and highlight the additional evaluation capabilities \sentbench provides.

\paragraph{Long-horizon evaluations.} Agents today still struggle with long-running tasks, as shown by METR's time-horizon work \citep{kwa2025metr}. Defining the 50\%-task-completion time horizon as the amount of human time required to complete tasks that an AI agent can solve with at least 50\% reliability, frontier models reached approximately fifty minutes in early 2025, continuing a trend of doubling roughly every seven months. Importantly, this horizon varies substantially by domain, including math, software, and visual computer-use tasks \citep{metr2025domains}. Follow-up analysis also finds a per-step hazard rate, showing that task success drops exponentially with task length \citep{ord2025halflife}.

RE-Bench \citep{wijk2024rebench} and HCAST \citep{rein2025hcast} are two foundational benchmarks for long-horizon tasks. Complementary benchmarks, such as \citep{wang2026horizon, motwani2026longcot, jang2026odysseys, garikaparthi2026can}, probe different axes of the long-horizon problem, including failure attribution, reasoning length, and duration estimation. Together, these works motivate \sentbench, since monitoring tasks often unfold over many minutes or hours, and can require many steps. However, whereas prior work largely measures tasks in which agents act continuously, \sentbench focuses on tasks where progress sometimes requires the agent to wait until the environment enters a state permitting forward progress.

\paragraph{Web and computer-use benchmarks.} 
A second line of related work introduces benchmarks for evaluating agents that operate web browsers or windowed operating systems. WebArena \citep{zhou2023webarena} and VisualWebArena \citep{koh2024visualwebarena} include realistic web applications, such as a forum, e-commerce platform, project management site, and content-editing environment, with more than 800 templated tasks. Similarly, WebVoyager \citep{he2024webvoyager} evaluates agents on fifteen consumer websites, Mind2Web \citep{deng2023mind2web} includes tasks across 137 sites, and AssistantBench \citep{yoran2024assistantbench} introduces 214 open-web research tasks that take humans a meaningful amount of time to complete. Building on this foundation, BrowserGym \citep{chezelles2024browsergym} combines several of these benchmarks into a single gym-like interface, while AgentBench \citep{liu2024agentbench} aggregates related tasks into a multi-domain suite. WebGames \citep{thomas2025webgames} adds interactive UI puzzles, and ST-WebAgentBench \citep{levy2024stwebagent} evaluates safety and trust dimensions. Finally, looking beyond web applications, OSWorld \citep{xie2024osworld} introduces OS-level tasks involving desktop applications. Like the long-horizon benchmarks discussed above, these benchmarks generally evaluate agents on tasks that run to completion in a single uninterrupted execution loop, with little or no waiting required.

\paragraph{Workplace and multi-application simulators.} A third line of work introduces benchmarks for multi-application workflows that more closely resemble digital-worker activity. AppWorld \citep{trivedi2024appworld} simulates nine applications and roughly one hundred users through API-style tool calls. TheAgentCompany \citep{xu2024agentcompany} simulates a small software company with internal websites and data; on this benchmark, the most competitive agent at release achieved approximately 30\% autonomous task completion. OdysseyBench \citep{wang2025odyssey} targets office-suite workflows spanning multiple documents and multi-session dialogues. AMA-Bench \citep{zhao2026amabench} measures long-horizon \emph{memory} for agentic applications. More recent multi-application benchmarks and simulators~\citep{li2026clawsbench, xiu2026astra, fu2026agent, lu2026alphaeval} broaden the scope to more applications, services, and longer tasks. Like \sentbench, these benchmarks either self-host realistic application interfaces or move toward production-grounded tasks~\citep{lu2026alphaeval}; however, they still score agents on contiguous execution. \sentbench complements this work by evaluating agents not only on how they act, but also on when they act.

\paragraph{Monitoring and scheduled agents.} Several products and systems have begun to address the need to run agents periodically, on a schedule, or to monitor an environment until specified conditions are met. In one early instance, \citet{thacker2024snooze} argued for a ``snooze button'' that would let an agent go dormant until an optimal time, similar to the \texttt{sleep} evaluated earlier. More recently, OpenAI's scheduled-task feature \citep{openai2025tasks} in ChatGPT \citep{openai2025chatgptagent}, along with a similar feature in Claude Cowork~\cite{anthropic2026claudecowork}, lets users schedule one-shot or recurring tasks. In \sentbench, however, events can happen at any time, so fixed scheduling approaches naturally degenerate into slow polling. \sentbench can be used directly to measure the effectiveness of such strategies, and you might expect to see slow reaction times.

More closely aligned with the agents considered in this paper, Yutori's Scouts \citep{yutori2025architecture} is a product marketed as ``always-on AI agents that monitor the Web for anything you care about.'' As with \sentbench, users describe scout tasks in natural language, and the service monitors web content on their behalf. Google \citep{google2026spark} also recently introduced both scheduled-task and monitoring capabilities in web-agent offerings, although the implementation details are not public.

Beyond web agents, Claude Code's \texttt{monitor} tool \citep{anthropic2025claudecodemonitor} lets Claude react to changes in terminal environments. This tool is conceptually similar to the \texttt{wait\_for} tool we evaluate in \sentbench, with the key difference being that \texttt{monitor} watches files, logs, and scripts for events, whereas \texttt{wait\_for} watches web application state. More importantly, these systems provide agents, while \sentbench provides a benchmark for measuring their effectiveness.

\paragraph{From scheduled events to environmental monitoring.} Across these categories, no benchmark makes waiting the primary task, although ARE~\cite{froger2025scaling} comes closest. ARE introduces a simulation platform in which time advances independently of the agent's actions, allowing scheduled events to fire in much the same way as in \sentbench. Within ARE, the most relevant benchmark is GAIA2~\cite{froger2026gaia2}, which consists of 1,120 scenarios across 12 smartphone app settings, including Messages, Contacts, and Shopping. A small subset of these tasks are temporal, testing an agent's ability to reason about and respond to time and events. Although frontier models reach a pass@1 of 42.1\% overall (GPT-5, High), they perform poorly on temporal tasks: GPT-5 (High) scores 0.0\%, while the best-performing model, Claude 4-Sonnet (thinking), reaches only 8.5\%. \sentbench shares ARE's premise that environments can evolve independently of agent actions, but differs in two key ways. First, GAIA2 is built around API access to applications, with a clear notification queue and event pipeline, whereas \sentbench requires agents to monitor natural, messy web pages. Second, GAIA2's temporal tasks often specify temporal constraints or expectations in the task prompt, whereas \sentbench tasks require agents to infer the triggering condition and waiting strategy.

Another close peer to \sentbench is Pare-Bench~\citep{nathani2026proactive}, which evaluates proactive assistance across 143 scenarios. In Pare-Bench, agents are scored on whether they intervene at the right moment for a simulated user. Both Pare-Bench and \sentbench evaluate wait-then-act scenarios, but they differ in their trigger sources and evolution mechanisms: Pare-Bench centers on user-simulator events, while \sentbench uses a broader range of environment events, such as a new song being released or a new paper being published.

In summary, \sentbench fills an important gap by benchmarking agents on their ability to monitor and respond to changes in web applications without relying on direct API access or dedicated notification channels. Table~\ref{tab:bench-comparison} summarizes this contrast.

\begin{table}[h]
\centering
\small
\begin{tabularx}{\textwidth}{@{}>{\raggedright\arraybackslash}p{5.2cm} >{\raggedright\arraybackslash}p{3.4cm} L@{}}
\toprule
\textbf{Benchmark} & \textbf{Scale} & \textbf{Long-horizon framing} \\
\midrule
GAIA \citep{mialon2023gaia} & 466 questions & Single-turn QA, no waiting \\
WebArena \citep{zhou2023webarena} & 812 tasks, 4 sites & Multi-step navigation, single session. \\
WebVoyager \citep{he2024webvoyager} & 643 tasks, 15 sites & Live-web navigation, single session. \\
Mind2Web \citep{deng2023mind2web} & 2350 tasks, 137 sites & Multi-step navigation, single session. \\
AssistantBench \citep{yoran2024assistantbench} & 214 tasks, 200+ sites & Auto-verifiable info-seeking, single session. \\
OSWorld \citep{xie2024osworld} & 369 tasks & OS-level control, single session. \\
AppWorld \citep{trivedi2024appworld} & 750 tasks, 9 apps & Multi-app API chains, single session. \\
TheAgentCompany \citep{xu2024agentcompany} & 175 tasks & Multi-app workflows, single session. \\
WebGames \citep{thomas2025webgames} & 50+ tasks & Interactive UI puzzles, single session. \\
OdysseyBench \citep{wang2025odyssey} & $\approx$602 tasks & Office workflows, multi-session dialogue. \\
Pare-Bench \citep{nathani2026proactive} & 143 tasks, 4 application domains & Asynchronous events as notifications, user simulator events \\ 
GAIA2 \citep{froger2026gaia2} & 1,120 tasks, 12 apps &  Asynchronous events as notifications, timing for triggers supplied in prompts \\
\textbf{\sentbench} (ours) & 100 scenarios, 10 envs & Externally-scheduled triggers discovered by polling, configurable-duration waits, plus no-op scenarios. \\
\bottomrule
\end{tabularx}
\caption{Public benchmarks for evaluating LLM-based agents on web and workplace tasks. \sentbench is the only benchmark whose tasks require the agent to wait for an externally-scheduled condition while performing periodic state checks.}
\label{tab:bench-comparison}
\end{table}

\section{Conclusion}
\label{sec:conclusion}

As AI agents take on longer-running tasks, they will increasingly encounter situations in which progress depends not on immediate action, but on recognizing when to wait, monitor, and respond at the right moment. In this paper, we introduced \sentbench, a benchmark for evaluating agents on time-evolving monitoring tasks across 10 synthetic web environments and 100 task scenarios. \sentbench measures not only whether agents complete these tasks, but also how quickly they react and how many resources they consume. Our baseline evaluations show that model choice matters, but that harness and tool design can matter just as much: a simple \texttt{wait\_for} tool substantially reduces cost relative to \texttt{sleep}, especially as task duration increases, while maintaining or improving task success in most conditions. To support reproducible evaluation, we release the code, task scenarios, synthetic catalogs, data-generation pipeline, and evaluation protocol at \url{https://github.com/microsoft/sentinel_environments}, providing a shared benchmark for studying how agents perform on monitoring tasks. More generally, these results and artifacts highlight the need to evaluate agents under conditions where the world changes independently of their actions, and to build agents that are more patient, efficient, and appropriately responsive.

\newpage

\bibliographystyle{plainnat}
\bibliography{references}

\newpage

\appendix

\newenvironment{promptquote}
  {\par\nobreak\vspace{-0.4em}%
   \begingroup
   \small\itshape
   \setlength{\leftskip}{1.8em}%
   \setlength{\rightskip}{1.8em}%
   \noindent``\ignorespaces}
  {\unskip''\par\endgroup}

\section{Data generation pipeline}
\label{app:datagen}

In this section, we detail the pipeline used in \sentbench to generate the synthetic data that populate the environments. We begin by motivating the need for such a process, then describe the procedures used to generate both multimedia and text content.

\subsection{Why Synthetic Environments?}
\label{app:why-synthetic}
Synthetic web environments are useful for evaluating agents for several important reasons, and they are especially important for evaluating agents on monitoring tasks. First, benchmark tasks do not arise from organic user needs. They are synthetically generated to resemble real user requests, but do not correspond to any specific person's immediate need. Directing benchmark traffic to live user-facing websites could place undue strain or costs on those sites, and might conflict with their posted terms of service. Operating in synthetic environments avoids these problems.

More importantly, synthetic environments are necessary for benchmark reproducibility, especially for monitoring tasks where timing is a critical factor. \sentbench needs events to unfold on a controlled schedule. For example, an email might arrive at minute~12, a stock might cross a threshold at minute~40, or a connection request might appear partway through a run and require a response. Live platforms such as Instagram or YouTube cannot provide this level of control, nor can facsimiles that track live data, such as a mock trading platform that follows market movements. In both cases, repeatable evaluation would be impossible. Self-contained clones backed by synthetic data give us a world that we can pause, script, and replay deterministically.

Finally, this same control makes these environments valuable beyond evaluation. The machinery that scripts a test scenario can also generate large volumes of labeled trajectories for \emph{training} monitoring agents, or environments for reinforcement learning.

\subsection{Synthetic multimedia content}
\label{app:datagen-text-multimedia}

Within this controlled world, a few design choices shape the data. Every asset is generated rather than scraped from the web, so the benchmark contains no real photos, names, or PII and can be released openly. To avoid introducing brand marks, we also try to keep generated media mostly free of rendered text. All media assets are generated media using open-weight models (Table~\ref{tab:datagen-models}), which keeps the pipeline reproducible on common hardware. Finally, each synthetic persona remains coherent across environments. The same likeness, biography, and interests follow a user from MicroDin to MicroChat to MicroGram, carried by a stable slug that threads through every stage of generation.

\begin{table}[H]
\centering
\small
\begin{tabularx}{\textwidth}{@{}l l L@{}}
\toprule
\textbf{Modality} & \textbf{Model} & \textbf{Settings} \\
\midrule
Images & FLUX.2-dev \citep{flux2dev} & 28 steps, guidance 4.0, bfloat16 \\
Videos & Wan2.2-T2V-14B \citep{wan2025} & 50 steps, 81 frames, 720p \\
Audio  & ACE-Step-HQ \citep{acestep2025} & 150 steps, 2 min, 44.1\,kHz \\
\bottomrule
\end{tabularx}
\caption{Three open-weights models cover the three modalities (Table~\ref{tab:datagen-models}). Generation runs on an NVIDIA B200 (183\,GB VRAM) node with CUDA 12.8. Approximate generation time: 100 images $\approx$ 15\,min; 50 videos $\approx$ 22\,hr; 100 music tracks $\approx$ 5\,hr.}
\label{tab:datagen-models}
\end{table}

\begin{figure}[t]
\centering
\includegraphics[width=\textwidth]{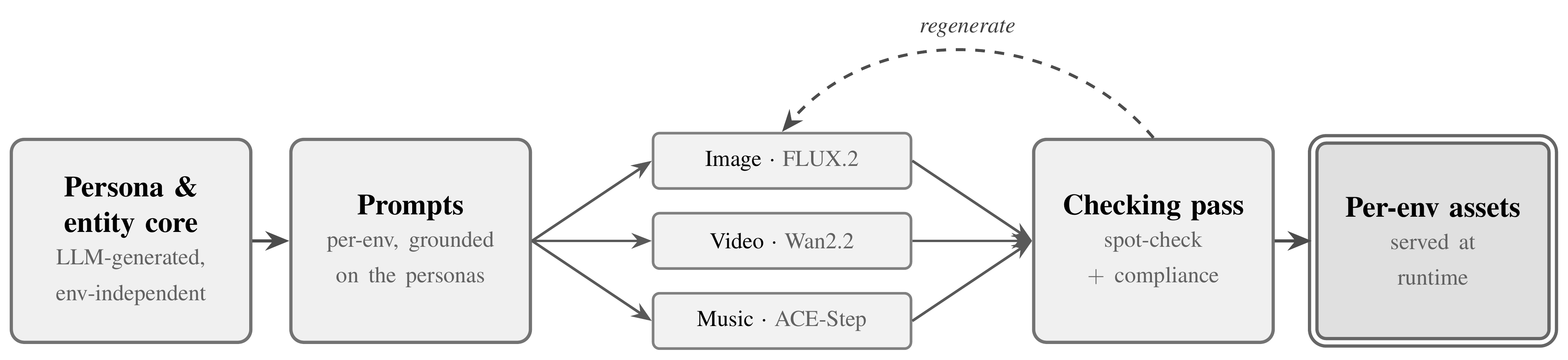}
\caption{The media-generation pipeline. An environment-independent core of LLM-generated personas and entities is shared across all ten environments. For each environment, a prompt is written for every image, video, and audio asset, grounded on the relevant personas. The prompts fan out to three generative models (image, video, audio). Generated media passes a checking pass (spot-check and compliance), which loops failures back for regeneration, before the per-environment media assets are assembled. Structured text catalogs are generated separately (Section~\ref{app:datagen-text}).}
\label{fig:datagen-pipeline}
\end{figure}

Figure~\ref{fig:datagen-pipeline} shows the flow for the visual and audio assets (the structured text catalogs are generated separately; see Section~\ref{app:datagen-text}). Generation begins with an environment-independent \emph{core}: a shared pool of LLM-generated personas and entities (companies, bands, channels, institutions) that is created once and reused across all ten environments to help with world coherence. For each environment we then write a \emph{prompt} for every image, video, and audio asset, grounded on the relevant personas and entities so the asset fits that environment and that identity; each prompt is tagged with a slug recording which persona, post, or team it belongs to, and prompts that could plausibly render text also carry anti-text guardrails. The prompts \emph{fan out to the appropriate generative model}, (images with FLUX.2-dev, video with Wan2.2, audio with ACE-Step). Generated media then pass an additional check that routes unusable assets (garbled text, accidental brand marks) back for regeneration. The result is the set of \emph{per-environment media assets} the apps serve at runtime. Table \ref{tab:datagen-assets} details the number of assets, by type, included in \sentbench.

\begin{table}[b]
\centering
\small
\begin{tabularx}{\textwidth}{@{}l >{\raggedright\arraybackslash}p{4.5cm} L@{}}
\toprule
\textbf{Environment} & \textbf{Description} & \textbf{Generated assets} \\
\midrule
Users        & Synthetic profiles      & 100 avatars, 100 banners \\
MicroDin     & Professional network    & 91 company logos, 91 banners, 20 posts \\
MicroMail    & Email                   & 25 document attachments \\
MicroTube    & Video platform          & 19 channels, 50 videos, 50 thumbnails \\
MicroFy      & Music streaming         & 100 tracks with covers \\
MicroGram    & Photo sharing           & 300 posts, 50 stories \\
MicroChat    & Team messaging          & 10 team icons \\
\bottomrule
\end{tabularx}
\caption{Asset counts per environment. The four text-only environments (MicroHub, MicroHood, MicroLendar, MicroScholar) use structured JSONL data with no AI-generated media.}
\label{tab:datagen-assets}
\end{table}

\subsection{Synthetic textual content}
\label{app:datagen-text}

All catalog text for the synthetic personas, including chat messages, emails, captions, post bodies, calendar event descriptions, and paper abstracts, is generated separately using ChatGPT 5.4. Avatars and banners are seeded from structured personas drawn from a 100-user synthetic identity table, with fields for name, age, gender, race, location, job title, biography, personality, and interests. Each persona's slug is reused across environment catalogs, so a single user retains a coherent identity across environments, including the same face, biography, and interests. A representative avatar prompt is shown below.

\begin{promptquote}
create a professional headshot for this person: name: Chris Taylor, age: 29, gender: male, race: White, location: San Francisco, CA, jobTitle: Product Associate, bio: A product-focused employee trying to build useful things. Juggling features, bugs, and way too many browser tabs., personality: adaptable, inquisitive, team-player, interests: tech, product development, agile methodologies, rock climbing
\end{promptquote}

Every prompt that could plausibly produce readable text includes explicit anti-text directives. Video prompts repeat ``no text no words no letters'' verbatim on each line. Audio prompts describe genre and mood without referencing real artists. Representative prompts follow:

\textit{MicroTube video (Wan2.2-T2V-14B):}
\begin{promptquote}
Sleek electric car racing on a futuristic track, dynamic camera angles, cinematic lighting, high-speed motion blur, professional automotive footage, no text no words no letters. 16:9 landscape.
\end{promptquote}

\textit{MicroGram post (FLUX.2-dev):}
\begin{promptquote}
Data science workshop presentation, whiteboard with ML diagrams, engaged audience, tech conference setting, square format 1:1, natural lighting, social media aesthetic.
\end{promptquote}

\textit{MicroFy track (ACE-Step-HQ):}
\begin{promptquote}
indie folk, acoustic guitar, warm vocals, gentle strumming, summer vibes, mellow and peaceful
\end{promptquote}

\textit{MicroChat team icon (FLUX.2-dev, abstract logo):}
\begin{promptquote}
Minimal vector icon for Engineering (Software Development): simplified circuit trace + code brackets ``$<>$'' motif, flat geometric monoline shapes with rounded corners, lots of negative space, pure white background, single accent color \#5B5FC7 (no gradients), centered icon, no text, no shadows, crisp edges.
\end{promptquote}

\section{Per-environment details}
\label{app:envs}

This appendix gives an overview of each of the 10 environments (Figures~\ref{fig:micromail_environment}-\ref{fig:microtube_environment}). For each environment we show the two most representative populated views side by side, with the real-world analog, the catalog scale, and the supported end-to-end actions in the caption.

\newcommand{\envfig}[7]{%
  \subsection{#2 \textemdash\ #3}%
  \begin{figure}[H]
    \centering
    \setlength{\fboxsep}{2pt}%
    \setlength{\fboxrule}{0.6pt}%
    \fbox{\includegraphics[width=0.478\textwidth]{figures/environments/#1/#4}}%
    \hfill
    \fbox{\includegraphics[width=0.478\textwidth]{figures/environments/#1/#5}}%
    \caption{#6}
    \label{#7}
  \end{figure}
}

\envfig{micromail}{MicroMail}{Email}{02_loaded.png}{04_view_b.png}{MicroMail, modeled on Gmail/Outlook. Left: the populated inbox with an email open in the reading pane, one of seven folders (Inbox, Sent, Drafts, Archive, Junk, Deleted, Scheduled) over a catalog of 260 emails with realistic subjects and bodies plus 25 attachments. Right: the compose surface (a new-message modal over the inbox). Supported end-to-end actions include mark-read, flag, move-to-folder, delete, restore, and client-side search; scenario events inject new emails over the run.}{fig:micromail_environment}

\envfig{microchat}{MicroChat}{Team messaging}{02_loaded.png}{04_view_b.png}{MicroChat, modeled on Slack/Teams. Left: the conversation list, with collapsible favorites and channels, unread-count badges, and presence indicators, backed by 200 messages across 30 conversations and 10 teams. Right: the calls tab, showing recent and missed calls and a speed-dial grid (20 calls in the catalog). Supported actions include mark-read, react, mute, and pin; events inject new messages and calls.}{fig:microchat_environment}

\envfig{microdin}{MicroDin}{Professional networking}{02_loaded.png}{03_view_a.png}{MicroDin, modeled on LinkedIn. Left: the home feed. Right: the My Network page, with pending invitations and connection suggestions. Catalog: 50 posts, 35 connections (some pending), 25 job listings, and 12 notifications. Supported end-to-end actions include liking a post, accepting or rejecting a connection, applying to a job, and marking a notification read; events inject new posts, jobs, connection requests, and notifications.}{fig:microdin_environment}

\envfig{microfy}{MicroFy}{Music streaming}{02_loaded.png}{04_view_b.png}{MicroFy, modeled on Spotify. Left: the music library track grid. Right: an artist/playlist page. Catalog: 100 tracks across 20 playlists, each with full lyrics in JSON and AI-generated cover art. Supported actions: like/unlike, create playlist, add to playlist, follow artist; events inject new releases and increment play and follower counts.}{fig:microfy_environment}

\envfig{microgram}{MicroGram}{Photo sharing}{02_loaded.png}{03_view_a.png}{MicroGram, modeled on Instagram. Left: a post in the feed. Right: the photo grid on a profile. Catalog: 300 posts with hashtagged captions, 50 stories, 35 activity items, and 6 DM conversations, all images AI-generated. Supported actions: like, save, follow, mark story viewed, mark activity read; events inject posts, stories, likes, and follows.}{fig:microgram_environment}

\envfig{microhood}{MicroHood}{Stock trading}{02_loaded.png}{04_view_b.png}{MicroHood, modeled on Robinhood. Left: the portfolio dashboard with its value chart. Right: the market view, with crypto and equity cards, a portfolio value chart, and a Buy/Sell order panel for the selected stock. Catalog: 50 stocks with realistic price, change, and percent fields, a 4-item watchlist, and 20 news items. Prices interpolate between authored waypoints in simulation time, so portfolio value drifts continuously; supported actions include placing market or limit buy/sell orders and editing the watchlist.}{fig:microhood_environment}

\envfig{microhub}{MicroHub}{Code hosting}{02_loaded.png}{03_view_a.png}{MicroHub, modeled on GitHub. Left: the repository overview. Right: the issues list. Catalog: one repository (with star, fork, and watch counts), 20 issues, 12 pull requests, 25 commits, 26 files, and 7 workflow runs. Supported actions: create issue, merge PR, star, fork, comment; events inject new issues, PRs, commits, and workflow runs, and the star count interpolates between waypoints.}{fig:microhub_environment}

\envfig{microlendar}{MicroLendar}{Calendar}{02_loaded.png}{04_view_b.png}{MicroLendar, modeled on Google Calendar. Left: the month grid. Right: the schedule/agenda view. Catalog: 57 events with realistic titles, times, locations, and attendees, plus a task list. Supported actions: create, edit, delete, and drag-to-reschedule events, and create or toggle tasks. The calendar surface is fixed after preload (no time-evolution).}{fig:microlendar_environment}

\envfig{microscholar}{MicroScholar}{Academic search}{03_view_a.png}{04_view_b.png}{MicroScholar, modeled on Google Scholar. Left: a search results list. Right: an author profile with citation history. Catalog: 100 papers with realistic titles, authors, years, and citation counts, 35 coauthor relationships, and 15 alerts. Supported actions: search, save/unsave, mark alert read, and export citations in five styles; events inject new papers and alerts and can increment citation counts.}{fig:microscholar_environment}

\envfig{microtube}{MicroTube}{Video streaming}{02_loaded.png}{04_view_b.png}{MicroTube, modeled on YouTube. Left: the subscriptions feed. Right: the video player with comments. Catalog: 50 videos, 20 channels, 7 comments, and 10 notifications. Supported actions: like or dislike, save to playlist, comment, subscribe, mark notification read; events inject new videos, comments, and notifications, and view and subscriber counts drift.}{fig:microtube_environment}

\section{SentinelBench Tasks}
\label{app:task_list}
Table~\ref{tab:all-prompts} presents a catalog of the \sentbench tasks, consisting of 100 tasks (10 per environment) with 20 no-operation tasks. Each task has a categorization along the axes of \textit{Action Requirement} (passive, active, or no-op) and \textit{Criterion Type} (absolute, or relative).

{\footnotesize
\setlength{\tabcolsep}{4pt}
\renewcommand{\arraystretch}{1.05}
\begin{longtable}{@{}p{0.85\linewidth} l@{}}
\caption{All 100 \sentbench task prompts, grouped by environment and sorted within each group by action requirement and criterion type.}\label{tab:all-prompts}\\
\toprule
\textbf{Task prompt} & \textbf{Categorization} \\
\midrule
\endfirsthead
\multicolumn{2}{l}{\textit{Table \ref{tab:all-prompts} continued from previous page}} \\
\toprule
\textbf{Task prompt} & \textbf{Categorization} \\
\midrule
\endhead
\midrule
\multicolumn{2}{r}{\textit{continued on next page}} \\
\endfoot
\bottomrule
\endlastfoot
\multicolumn{2}{@{}l}{\textbf{MicroChat}} \\
\addlinespace[2pt]
``Keep an eye on my calls tab and let me know when I've missed at least 3 calls.'' & passive $\cdot$ absolute \\
``Keep an eye on my chats and let me know when I have 10 unread messages.'' & passive $\cdot$ absolute \\
``I've already missed a couple of calls this morning. Let me know when 3 more missed calls come in.'' & passive $\cdot$ relative \\
``Let me know when unread messages show up in 3 more of my chats than right now. Just watch the sidebar---don't open any of the conversations, I don't want people seeing 'seen' on their messages before I've had a chance to read them.'' & passive $\cdot$ relative \\
``Let me know the moment someone sends me a new file in any of my chats. Attachments aren't visible from the sidebar, so you'll have to keep an eye on the various chats. I want to be sure I don't miss any important files that come in.'' & active $\cdot$ absolute \\
``Keep an eye on my chat with Alex Chen. If he reports any problems with payment processing, read the message to get the full scope of the problem, then let me know right away.'' & active $\cdot$ absolute \\
``I'm getting a lot of @mentions today. As new messages come in, read them in full, then let me know as soon as I have 3 more with mentions.'' & active $\cdot$ relative \\
``I've already got a couple of urgent pings this morning. Let me know when at least 3 more urgent messages arrive. Note: urgent messages are only marked as such inside the conversation, not in the sidebar.'' & active $\cdot$ relative \\
``Let me know the moment Charles Davis sends me a file in any of my chats. You'll need to open each conversation to check---attachments aren't visible from the sidebar.'' & no-op \\
``Let me know when Diana Miller @mentions me in any of my chats. Note: you'll need to open conversations to see which messages are mentions.'' & no-op \\
\addlinespace[8pt]
\multicolumn{2}{@{}l}{\textbf{MicroDin}} \\
\addlinespace[2pt]
``Keep an eye on my network page and tell me when I have at least 15 pending connection requests.'' & passive $\cdot$ absolute \\
``Keep an eye on the jobs page and tell me when there are 8 or more job listings.'' & passive $\cdot$ absolute \\
``Ping me once at least 3 new unread notifications come in.'' & passive $\cdot$ relative \\
``Keep the feed open and tell me once at least 5 new posts have shown up since I left.'' & passive $\cdot$ relative \\
``Watch the jobs page. If a new role shows up that requires API documentation experience, apply to it for me.'' & active $\cdot$ absolute \\
``Keep an eye on the jobs page. If a new role mentions Kubernetes in the requirements, apply to it and let me know.'' & active $\cdot$ absolute \\
``Two new Python-heavy roles should come through today. Apply to both when you see them---you'll need to open the job details to check the requirements.'' & active $\cdot$ relative \\
``If any new jobs pop up that ask for React experience, apply to at least two of them for me. You'll need to open each job listing to check the requirements.'' & active $\cdot$ relative \\
``If any new jobs pop up that ask for COBOL experience, apply to at least two of them for me. You'll need to open each job listing to check the requirements.'' & no-op \\
``Keep an eye on the jobs page and tell me when there are 50 or more job listings.'' & no-op \\
\addlinespace[8pt]
\multicolumn{2}{@{}l}{\textbf{MicroFy}} \\
\addlinespace[2pt]
``MicroFy is rolling out playlist updates today. Once you've seen at least 3 new playlists appear in the sidebar, let me know.'' & passive $\cdot$ relative \\
``Follow the next 4 artists to appear in the trending feed. To follow an artist, you must open their Artist page.'' & active $\cdot$ absolute \\
``Keep checking for new trending songs. Like the next 4 new songs to appear.'' & active $\cdot$ absolute \\
``Watch the trending feed. When a song drops whose lyrics mention 'subway', like it for me.'' & active $\cdot$ absolute \\
``Keep checking for new songs in trending. Play the next 3 new songs to appear.'' & active $\cdot$ absolute \\
``I'm collecting songs that mention 'golden' in their lyrics. Watch for new trending songs that fit---when two more come in, play both to confirm they match.'' & active $\cdot$ relative \\
``I'm watching for a new blues track. When one of the new trending songs has 'whiskey' in the lyrics, like it.'' & active $\cdot$ relative \\
``Keep an eye on the trending feed today---once at least 5 new tracks have dropped since I logged in, save them all to my Liked Songs.'' & active $\cdot$ relative \\
``I've been hearing buzz that Synthwave Dreams has a release dropping today. Watch the trending feed---when one of their tracks shows up, open their Artist page and follow them.'' & no-op \\
``Watch the trending feed. When a song drops whose lyrics mention 'tangerine', like it for me.'' & no-op \\
\addlinespace[8pt]
\multicolumn{2}{@{}l}{\textbf{MicroGram}} \\
\addlinespace[2pt]
``Let me know when any post in my feed gets 4 or more comments.'' & passive $\cdot$ absolute \\
``Keep an eye on the story bar. Let me know when 3 new stories appear.'' & passive $\cdot$ absolute \\
``I've already got some notifications in my activity feed---keep an eye on it and let me know when 4 more show up.'' & passive $\cdot$ relative \\
``I already have a bunch of posts in my feed---let me know when 5 more new posts show up.'' & passive $\cdot$ relative \\
``Watch my DMs for a new message about a mockup---the word might not show in the preview, so you'll need to open conversations to check.'' & active $\cdot$ absolute \\
``For the next 3 new creators to appear in my feed, open their profiles and follow them.'' & active $\cdot$ absolute \\
``Keep an eye on my feed and like the next 3 new posts that show up.'' & active $\cdot$ absolute \\
``I've noticed @ahall has been commenting on posts in my feed---keep checking new posts as they come in, and let me know when 3 more posts show up that have a comment from @ahall (you'll need to open each post to see who commented).'' & active $\cdot$ relative \\
``I've noticed @dmiller has been commenting on posts in my feed---keep checking new posts as they come in, and let me know when 3 more posts show up that have a comment from @dmiller (you'll need to open each post to see who commented).'' & no-op \\
``Watch my DMs for a new message about a merger---the word might not show in the preview, so you'll need to open conversations to check.'' & no-op \\
\addlinespace[8pt]
\multicolumn{2}{@{}l}{\textbf{MicroHood}} \\
\addlinespace[2pt]
``Keep an eye on my portfolio and tell me when it reaches \$108,837.'' & passive $\cdot$ absolute \\
``Watch ChipForge Industries (CHIP) for me and let me know when it reaches at least \$520.'' & passive $\cdot$ absolute \\
``Keep an eye on the news feed in MicroHood and let me know when 3 new articles have appeared since I opened the page.'' & passive $\cdot$ relative \\
``Watch my portfolio value and let me know when it's up 5\% from where it started.'' & passive $\cdot$ relative \\
``Keep an eye on DRNE for me. If it hits \$42.20 or higher, place a market order to buy 2 shares and let me know once it's done.'' & active $\cdot$ absolute \\
``Watch MCRO for me. If it goes above \$480, sell 5 shares at market price and let me know.'' & active $\cdot$ absolute \\
``Watch Voltaic Motors (VOLT) for me. If it drops more than 10\% from its current price, buy 3 shares at market price and let me know.'' & active $\cdot$ relative \\
``Keep an eye on Neos Pharmaceuticals (NEOS). If it goes up 15\% or more from where it is now, sell all 12 of my shares to lock in the gains.'' & active $\cdot$ relative \\
``Keep an eye on AgriTech Innovations (AGRI) for me. If it hits \$500 or higher, place a market order to buy 5 shares and let me know once it's done.'' & no-op \\
``Watch Lithium Energy (LITH) for me and let me know when it reaches at least \$500.'' & no-op \\
\addlinespace[8pt]
\multicolumn{2}{@{}l}{\textbf{MicroHub}} \\
\addlinespace[2pt]
``Keep an eye on the Issues tab and tell me when there are 8 open issues.'' & passive $\cdot$ absolute \\
``Watch the star count on this repo and let me know when it hits at least 2,000.'' & passive $\cdot$ absolute \\
``I already have a few open issues. Let me know when 4 more get filed.'' & passive $\cdot$ relative \\
``The repo is getting some attention lately. Watch the star count and tell me when it goes up by 200 from where it is now.'' & passive $\cdot$ relative \\
``Watch for new issues. If one comes in that mentions compliance in its description, leave a comment on it and let me know.'' & active $\cdot$ absolute \\
``Keep an eye on new issues. If one shows up that mentions TOTP as a supported 2FA method, leave a comment on it and let me know.'' & active $\cdot$ absolute \\
``One of the existing issues has detailed steps to reproduce the bug. Watch for new issues, and let me know if another one comes in that also includes steps to reproduce---you'll need to read the issue descriptions to find it.'' & active $\cdot$ relative \\
``I saw a bug report earlier about the UI being unresponsive. Keep an eye on new issues and let me know if another one comes in that also mentions the UI being unresponsive---you'll need to read through the issue descriptions to check.'' & active $\cdot$ relative \\
``Watch for new issues. If one comes in that mentions telemetry in its description, leave a comment on it and let me know.'' & no-op \\
``Watch the star count on this repo and let me know when it hits at least 100,000.'' & no-op \\
\addlinespace[8pt]
\multicolumn{2}{@{}l}{\textbf{MicroLendar}} \\
\addlinespace[2pt]
``Keep an eye on my calendar and tell me when it reaches 21 or more events this month.'' & passive $\cdot$ absolute \\
``Keep an eye on my tasks and tell me when the list reaches at least 10 items.'' & passive $\cdot$ absolute \\
``I have a few personal events on my calendar already---keep an eye out and let me know when 2 more personal events get added.'' & passive $\cdot$ relative \\
``I already have a few things scheduled for today---let me know when 3 more events get added to today's schedule.'' & passive $\cdot$ relative \\
``I already have a few work events on my calendar. Tell me when 4 more work events get added in the next two weeks.'' & passive $\cdot$ relative \\
``I'm expecting a calendar invite about a trip to Hawaii---the details should be in the event description, so you'll need to open each event to check.'' & active $\cdot$ absolute \\
``I'm waiting for a meeting to be scheduled about the notifications feature---the specifics should be in the event description, not the title, so you'll need to open each event to check.'' & active $\cdot$ absolute \\
``I have an event coming up that mentions analytics in its description---let me know if another event gets added that also mentions analytics in its description, since you'll need to open each event to check.'' & active $\cdot$ relative \\
``I'm expecting a calendar invite about a trip to Iceland---the details should be in the event description, so you'll need to open each event to check.'' & no-op \\
``Keep an eye on my calendar and tell me when it reaches 100 or more events this month.'' & no-op \\
\addlinespace[8pt]
\multicolumn{2}{@{}l}{\textbf{MicroMail}} \\
\addlinespace[2pt]
``Let me know when I get an email from Kevin Lee.'' & passive $\cdot$ absolute \\
``Tell me when I have 10 or more unread emails.'' & passive $\cdot$ absolute \\
``Let me know when 5 new emails appear in my inbox (not counting any unread emails already there).'' & passive $\cdot$ relative \\
``Keep an eye on the junk folder for me, tell me when 3 more emails end up there.'' & passive $\cdot$ relative \\
``Watch for an email with the mobile app mockups attached.'' & active $\cdot$ absolute \\
``I'm expecting an email about platform scalability. Let me know as soon as it arrives. Note that 'platform scalability' might not be obvious from the subject, so you'll need to read through the emails.'' & active $\cdot$ absolute \\
``I got an email earlier mentioning a December deadline. Let me know if any new email arrives that also mentions the month of December. Note: December might only appear in the email body, so you should check there as well.'' & active $\cdot$ relative \\
``I'm expecting a few more emails where I'm CC'd. Let me know when 3 more come in.'' & active $\cdot$ relative \\
``Watch for an email with the press release attached.'' & no-op \\
``I'm expecting an important email from Lena Whitford. Let me know as soon as it arrives.'' & no-op \\
\addlinespace[8pt]
\multicolumn{2}{@{}l}{\textbf{MicroScholar}} \\
\addlinespace[2pt]
``Keep an eye on my alerts and let me know if one comes in about large language models.'' & passive $\cdot$ absolute \\
``I'm expecting the paper 'Autonomous AI Agents for Complex Task Execution' to be indexed soon. Please let me know as soon as it shows up in the search results.'' & passive $\cdot$ absolute \\
``I've got a couple of unread alerts already---let me know when 3 more unread alerts come in.'' & passive $\cdot$ relative \\
``I already have a bunch of papers recommended to me---keep an eye out and let me know when 5 more show up.'' & passive $\cdot$ relative \\
``I need to cite the paper 'Autonomous AI Agents for Complex Task Execution', but it may not have been indexed yet. Keep searching for it, and let me know as soon as you have the MLA citation.'' & active $\cdot$ absolute \\
``I'm looking for a paper that mentions 'downstream tasks' in its abstract---you'll need to read through the abstracts of new papers as they come in. When one does, save it to my library, then let me know.'' & active $\cdot$ absolute \\
``I'm tracking papers that mention 'unstructured environments'. I've already saved one to my library. Wait for, and save, 2 more as they are published.'' & active $\cdot$ relative \\
``Some papers by Deborah Jackson should be getting indexed soon---save each one to your library as it appears, and let me know once you've saved 3 of her papers.'' & active $\cdot$ relative \\
``Keep an eye on my alerts and let me know if one comes in about cold fusion.'' & no-op \\
``I'm expecting the paper 'Quantum Gravity in Discrete Spacetime' to be indexed soon. Please let me know as soon as it shows up in the search results.'' & no-op \\
\addlinespace[8pt]
\multicolumn{2}{@{}l}{\textbf{MicroTube}} \\
\addlinespace[2pt]
``Watch my notifications for me and let me know when I have at least 8.'' & passive $\cdot$ absolute \\
``Keep the Subscriptions feed open and tell me when there are 3 new videos from channels I already follow.'' & passive $\cdot$ absolute \\
``I already have a few notifications. Let me know when 4 more come in.'' & passive $\cdot$ relative \\
``I can see a few videos on the home page already. Let me know when 5 more new ones have shown up.'' & passive $\cdot$ relative \\
``A new Civilization game was just released---watch for new comments and let me know when you spot any mentioning Civilization.'' & active $\cdot$ absolute \\
``Keep an eye on Science Explained for me. If they post a new video, open it and like it.'' & active $\cdot$ absolute \\
``I saw a few comments on the gaming videos earlier. Keep checking and let me know when 3 more comments have been posted on them.'' & active $\cdot$ relative \\
``I already watched a couple videos from my subscriptions earlier. Keep watching that feed and watch 3 more new uploads for me as they appear.'' & active $\cdot$ relative \\
``I heard someone left a comment about Minecraft on one of the gaming videos---watch for new comments and let me know when you spot it.'' & no-op \\
``Keep an eye on Cinema Central for me. If they post a new video, open it and like it.'' & no-op \\
\end{longtable}
}

\clearpage
\section{The \texttt{wait\_for} tool}
\label{app:wait_for}

Section~\ref{sec:evaluation} describes the \texttt{wait\_for(condition, timeout)} tool at a high level. Algorithm~\ref{alg:wait_for} fills in the details that the prose elides: the adaptive intervals that govern how often the LLM is consulted and how often the page is reloaded, the deduplication of diff blocks the LLM has already seen, and the forced final check that runs after the timeout expires.

\begin{algorithm}[H]
\small
\caption{The \texttt{wait\_for(condition, timeout)} tool. The agent supplies a natural-language \emph{condition} and a maximum blocking \emph{timeout} (seconds); the tool returns a flag indicating whether the condition was met, along with the LLM's explanation.}
\label{alg:wait_for}
\begin{algorithmic}[1]
\Function{WaitFor}{$condition,\ timeout$}
  \State $start \gets \Call{Now}{}$;\ \Comment{timestamp when \Call{WaitFor}{} was called}
  \State $base \gets \Call{PageMarkdown}{}$ \Comment{baseline snapshot of the page}
  \State $reloadInt \gets 180$;\ \Comment{adaptive reload interval}
  \State $nextReload \gets start + reloadInt$ \Comment{schedule the next reload}
  \State $checkInt \gets 10$;\ \Comment{adaptive LLM-call rate limit}
  \State $lastLLMCheck \gets -\infty$;\  \Comment{time when the LLM was last called}
  \State $shownDiffs \gets \emptyset$;\ \Comment{Diffs already checked by the LLM}
  \While{$\Call{Now}{} - start < timeout$}
    \If{$\Call{Now}{} \geq nextReload$} \Comment{periodic reload guards against stale static pages}
      \State \Call{Reload}{}
      \State $reloadInt \gets \min(2 \cdot reloadInt,\ 480)$
      \State $nextReload \gets \Call{Now}{} + reloadInt$
    \EndIf
    \State $cur \gets \Call{PageMarkdown}{}$
    \State $blocks \gets \Call{UnifiedDiffBlocks}{base,\ cur}$ \Comment{contiguous changed regions}
    \State $newDiffs \gets blocks \setminus shownDiffs$ \Comment{drop blocks already evaluated by the LLM}
    \State $due \gets (newDiffs \neq \emptyset) \land (\Call{Now}{} - lastLLMCheck \geq checkInt)$
    \If{$due$}
      \State $(met,\ why) \gets \Call{AskLLM}{condition,\ newDiffs}$      
      \State $lastLLMCheck \gets \Call{Now}{}$;\
      \State $shownDiffs \gets shownDiffs \cup newDiffs$
      \If{$met$}
        \State \Return $(\textsc{true},\ why)$
      \EndIf
      \State $checkInt \gets \min(2 \cdot checkInt,\ 60)$ \Comment{back off on busy pages}
    \EndIf
    \State \Call{Sleep}{1}
  \EndWhile
  \Statex
  \State \Call{Reload}{} \Comment{do one final check after timeout with all diffs}
  \State $cur \gets \Call{PageMarkdown}{}$
  \State $blocks \gets \Call{UnifiedDiffBlocks}{base,\ cur}$
  \If{$blocks = \emptyset$}
    \State \Return $(\textsc{false},\ \text{``no changes detected before the timeout''})$
  \EndIf
  \State $(met,\ why) \gets \Call{AskLLM}{condition,\ blocks}$
  \State \Return $(met,\ why)$
\EndFunction
\end{algorithmic}
\end{algorithm}

\end{document}